# Complexity of Prioritized Default Logics

**Jussi Rintanen**                                    RINTANEN@INFORMATIK.UNI-ULM.DE
*Universität Ulm*
*Fakultät für Informatik*
*Albert-Einstein-Allee*
*89069 Ulm, GERMANY*

## Abstract

In default reasoning, usually not all possible ways of resolving conflicts between default rules are acceptable. Criteria expressing acceptable ways of resolving the conflicts may be hardwired in the inference mechanism, for example specificity in inheritance reasoning can be handled this way, or they may be given abstractly as an ordering on the default rules. In this article we investigate formalizations of the latter approach in Reiter's default logic. Our goal is to analyze and compare the computational properties of three such formalizations in terms of their computational complexity: the prioritized default logics of Baader and Hollunder, and Brewka, and a prioritized default logic that is based on lexicographic comparison. The analysis locates the propositional variants of these logics on the second and third levels of the polynomial hierarchy, and identifies the boundary between tractable and intractable inference for restricted classes of prioritized default theories.

## 1. Introduction

Nonmonotonic logics and related systems for nonmonotonic and default reasoning (Reiter, 1980; Moore, 1985; McCarthy, 1980) were developed for representing knowledge and forms of reasoning that are not conveniently expressible in monotonic logics, like the first-order predicate logic or propositional logics. In nonmonotonic logics inferences can be made on the basis of what cannot be inferred from a set of facts. When extending this set, some of the inferences may become invalid, and hence the set of inferable facts does not monotonically increase. For example, in the kind of reasoning expressed as inheritance networks (Horty, 1994), one network link may say that priests imbibe non-alcoholic beverages only. This is, in the absence of contrary information, a sufficient reason to conclude that a certain priest will not drink vodka. Without contradiction, information stating that the priest does drink vodka can be added, which retracts the previous conclusion.

The need to incorporate priority information to nonmonotonic logics (Lifschitz, 1985; Brewka, 1989; Geffner & Pearl, 1992; Ryan, 1992; Brewka, 1994; Baader & Hollunder, 1995) stems from the possibility that two default rules are in conflict. One source of such priority information is the specificity of defaults. When one rule says that priests usually do not drink and another says that men usually do, inferences concerning male priests should be based on the first one because it is more specific, as male priests are a small subset of men. Specificity as a basis for resolving conflicts between defaults has been investigated in the framework of path-based inheritance theories (Horty, 1994). In general, however, priorities may come from different sources, and therefore it is justified to investigate nonmonotonic reasoning with an abstract notion of priorities as orderings on defaults. In this setting the problem is to define what are the correct inferences in the presence of priorities. There have been many proposals of differing generality. Preferred subtheories (Brewka, 1989) and ordered theory presentations (Ryan, 1992) do not have as general a notion of defaults as Reiter's default logic, and they can both be translated to prerequisite-free normal default theories of a prioritized default logic (Rintanen, 1999). Also the definitions of model minimization in conditional entailment (Geffner & Pearl, 1992) and in prioritized circumscription (Lifschitz, 1985) do not directly support defaults with prerequisites.

In this work we concentrate on Brewka's (1994) and Baader and Hollunder's (1995) proposals for incorporating priorities to default logic, as well as a proposal that uses lexicographic comparison (Rintanen, 1999). These three proposals address default reasoning with defaults that have prerequisites. Like earlier work on complexity of nonmonotonic logics (Kautz & Selman, 1991; Stillman, 1990; Gottlob, 1992), the purpose of our investigation is to point out fundamental differences and similarities between these logics, characterized by polynomial time translatability between their decision problems, and to identify restricted classes of default theories where reasoning is tractable. The first kind of results are useful for example when developing theorem-proving techniques for the logics in question. The inexistence of polynomial time translations between two decision problems indicates that the techniques needed are likely to be different: it is not feasible to solve one decision problem by simply translating it to the other. Also, if polynomial time translations exist and they turn out to be simple, there is often no reason to treat the decision problems separately. The second kind of results, identification of tractable cases, directly give efficient, that is polynomial time, decision procedures for prioritized reasoning in special cases. The utility of these algorithms depends on the application at hand. In many cases the restrictions that lead to polynomial time decision procedures are too severe to make the procedures practically useful. Even though nonmonotonic reasoning usually cannot be performed in polynomial time, new developments in implementation techniques have made it possible to solve problems that previously were too difficult (Niemelä & Simons, 1996). Hence there are some prospects of making nonmonotonic reasoning practically useful, which makes the problem of introducing priority information in nonmonotonic reasoning more acute. This paper can be seen as giving guidelines in that direction.

The decision problems of propositional default logic are located on the second level of the polynomial hierarchy (Gottlob, 1992), and hence they do not belong to the classes NP or co-NP unless the polynomial hierarchy collapses to its first level. The proof of this result suggests that reasoning in default logic cannot be performed in polynomial time simply by restricting the formulae in default theories to a tractable subclass of propositional logic like 2-literal clauses (Even, Itai, & Shamir, 1976) or Horn clauses (Dowling & Gallier, 1984). This fact can be seen as a consequence of the possibility of conflicting defaults. A conflict between defaults gives rise to multiple extensions because a case analysis on alternative ways of resolving the conflict is required. For a default theory of size $n$, the number of conflicts may be proportional to $n$, and each conflict may double the number of extensions. Hence the number of extensions can be proportional to $2^n$. Priorities in many cases uniquely determine how a conflict between defaults is resolved, and hence the case analyses that lead to an exponential number of extensions can be avoided. This gives rise to the question whether priorities would in some cases produce a computational advantage in the sense that the decision problems could be solved more efficiently than in the unprioritized case.

To investigate these questions we consider three definitions of priorities in the framework of Reiter's default logic.

First we analyze the computational complexity of two closely related prioritized default logics (Brewka, 1994; Baader & Hollunder, 1995). These logics are based on the semiconstructive definition of extensions of default logic (Reiter, 1980): the priorities control the construction of extensions, ruling out those that do not respect the priorities. In the general case the complexity of the decision problems of these logics equals the complexity of those of Reiter's default logic, being complete for the second level of the polynomial hierarchy. When the priorities are a total ordering on the defaults, the complexity decreases by one level, leading to many tractable cases when the propositional reasoning involved is tractable, for example with Horn clauses or 2-literal clauses. With arbitrary strict partial orders there is no similar decrease.

We continue by analyzing a prioritized default logic that is based on comparing the Reiter extensions of a default theory by lexicographic comparison (Rintanen, 1999). The decision problem of this logic is harder than the decision problems of Reiter's default logic, assuming that the polynomial hierarchy does not collapse. For default theories with a total priority relation some syntactically restricted classes are easier than the corresponding unprioritized ones, but in general even total priorities do not reduce the complexity of the decision problems. For partial priorities tractability can be achieved only with extreme syntactic restrictions, the complexity being the same as with the prioritized default logics by Brewka and by Baader and Hollunder.

## 2. Preliminaries on Computational Complexity

In this section we introduce some basic concepts in computational complexity. For details see (Balcázar, Díaz, & Gabarró, 1995). The complexity class P consists of decision problems that are solvable in polynomial time by a deterministic Turing machine. NP is the class of decision problems that are solvable in polynomial time by a nondeterministic Turing machine. The class co-NP consists of problems the complements of which are in NP. In general, the class co-C consists of problems the complements of which are in the class C. The polynomial hierarchy PH is an infinite hierarchy of complexity classes $\Sigma_i^p$, $\Pi_i^p$ and $\Delta_i^p$ for all $i \geq 0$ that is defined by using oracle Turing machines as follows.

$$\begin{array}{ccccccc}
\Sigma_0^p & = & P & \Pi_0^p & = & P & \Delta_0^p & = & P \\
\Sigma_{i+1}^p & = & \mathrm{NP}^{\Sigma_i^p} & \Pi_{i+1}^p & = & \text{co-}\Sigma_{i+1}^p & \Delta_{i+1}^p & = & P^{\Sigma_i^p}
\end{array}$$

$C_1^{C_2}$ denotes the class of problems that is defined like the class $C_1$ except that oracle Turing machines that use an oracle for a problem in $C_2$ are used instead of Turing machines without such an oracle. Oracle Turing machines with an oracle for a problem $B$ are like ordinary Turing machines except that they may perform tests for membership in $B$ with constant cost. A problem $L$ is *Turing reducible* to a problem $L'$ if there is an oracle Turing machine with an oracle for $L'$ that solves $L$. The problem is Turing reducible in polynomial time if the oracle Turing machine solves $L$ with a polynomial number of execution steps. A problem $L$ is *C-hard* for a complexity class $C$ if all problems in $C$ are polynomial time *many-one reducible* to it; that is, for all problems $L' \in C$ there is a function $f_{L'}$ that can be computed in polynomial time on the size of its input and $f_{L'}(x) \in L$ if and only if $x \in L'$. We say

425

that the function $f_{L'}$ is a translation from $L'$ to $L$. A problem is $C$-*complete* if it belongs to the class $C$ and is $C$-hard.

## 3. Preliminaries on Default Logic

Default logic is one of the main formalizations of nonmonotonic reasoning (Reiter, 1980). A *default theory* $\Delta = \langle D, W \rangle$ consists of a set of *default rules* $\alpha{:}\beta_1, \ldots, \beta_n/\gamma$ where $\alpha$ (the *precondition*), $\gamma$ (the *conclusion*) and $\beta_i, i \in \{1, \ldots, n\}$ (the *justifications*) are formulae of the classical propositional logic, and a set $W$ of *objective facts* that also are formulae of the classical propositional logic. A default rule $\alpha{:}\beta_1, \ldots, \beta_n/\gamma$ can be used for inferring the fact $\gamma$ if $\alpha$ has been derived, and none of the formulae $\neg\beta_1, \ldots, \neg\beta_n$ can be derived. As derivability and underivability are mutually dependent, there is circularity in the definition of what is derivable in default logic. Unlike in monotonic logics, where the consequences of a set of formulae is defined as the formulae that can be derived by using the axioms of the logic and the inference rules, the conclusions of a default theory $\Delta = \langle D, W \rangle$ are defined as fixpoints of a nonmonotonic operator. The operator may have several fixpoints, none of which is the least fixpoint, and different fixpoints can be seen as a result of applying different mutually incompatible sets of default rules. The fixpoints of the operator for a default theory are the *extensions* of the default theory.

Informally, the construction of each extension of a default theory starts from the objective facts $W$, and proceeds by adding conclusions of default rules the preconditions of which have already been derived, and the justifications of which have not been and will not later be contradicted. The construction ends when no more defaults can be applied.

**Example 3.1** Define $\Delta = \langle D, W \rangle$ by

$$D = \left\{ \frac{priest : \neg drinks\text{-}vodka}{\neg drinks\text{-}vodka}, \frac{man : drinks\text{-}vodka}{drinks\text{-}vodka}, \frac{priest : man}{man} \right\}, \text{ and}$$
$$W = \{priest\}.$$

A default $\alpha{:}\beta/\beta$ can be interpreted as saying that an individual who has property $\alpha$, can be assumed to also have the property $\beta$ if this is consistent with what else is known. The default theory $\Delta$ has two extensions, $E = Cn(\{priest, man, \neg drinks\text{-}vodka\})$ and $E' = Cn(\{priest, man, drinks\text{-}vodka\})$, that represent the two possibilities of resolving the conflict between the first two defaults in $D$. The extension $E$ corresponds to the choice to apply the first default, and the extension $E'$ to the choice to apply the second. □

The fixpoint definition of extensions is given next. The language of the propositional logic is denoted by $\mathcal{L}$. The closure of a set of formulae $S$ under logical consequence is $Cn(S) = \{\phi \in \mathcal{L}|S \models \phi\}$. The set $\{\alpha{:}\beta_1, \ldots, \beta_n/\gamma | n \geq 0, \{\alpha, \beta_1, \ldots, \beta_n, \gamma\} \subseteq \mathcal{L}\}$ of default rules is denoted by $\mathcal{D}$. The *size* of a default theory $\langle D, W \rangle$ is the sum of the lengths of the formulae in $W$ and in defaults in $D$. A default theory is finite if $D$ and $W$ are finite.

**Definition 3.1 (Reiter, 1980)** *Let* $\Delta = \langle D, W \rangle$ *be a default theory. For any set of formulae* $S \subseteq \mathcal{L}$, *let* $\Gamma(S)$ *be the smallest set such that* $W \subseteq \Gamma(S)$, $Cn(\Gamma(S)) = \Gamma(S)$, *and if* $\alpha{:}\beta_1, \ldots, \beta_n/\gamma \in D$ *and* $\alpha \in \Gamma(S)$ *and* $\{\neg\beta_1, \ldots, \neg\beta_n\} \cap S = \emptyset$, *then* $\gamma \in \Gamma(S)$. *A set of formulae* $E \subseteq \mathcal{L}$ *is an extension for* $\Delta$ *if and only if* $\Gamma(E) = E$.

More procedural is the *semiconstructive* definition, so called because it suggests a nondeterministic procedure for computing extensions.

**Theorem 3.2 (Reiter, 1980)** *Let $E \subseteq \mathcal{L}$ be a set of formulae, and let $\Delta = \langle D, W \rangle$ be a default theory. Define $E_0 = W$ and for all $i \geq 0$, $E_{i+1} = Cn(E_i) \cup \{\gamma | \alpha{:}\beta_1, \ldots, \beta_n/\gamma \in D, \alpha \in E_i, \{\neg\beta_1, \ldots, \neg\beta_n\} \cap E = \emptyset\}$. Then $E$ is an extension of $\Delta$ if and only if $E = \bigcup_{i \geq 0} E_i$.*

The set of *generating defaults* of an extension identifies the extension uniquely, and is of finite size whenever the number of defaults is finite.

**Definition 3.3 (Reiter, 1980)** *Suppose $\Delta = \langle D, W \rangle$ is a default theory and $E$ is an extension of $\Delta$. The set of* generating defaults *of $E$ with respect to $\Delta$ is*

$$GD(E, \Delta) = \left\{ \frac{\alpha : \beta_1, \ldots, \beta_n}{\gamma} \in D \,\middle|\, \alpha \in E \text{ and } \{\neg\beta_1, \ldots, \neg\beta_n\} \cap E = \emptyset \right\}.$$

**Theorem 3.4 (Reiter, 1980)** *Suppose $E$ is an extension of a default theory $\Delta = \langle D, W \rangle$. Then $E = Cn(W \cup \{\gamma | \alpha{:}\beta_1, \ldots, \beta_n/\gamma \in GD(E, \Delta)\})$.*

The standard consequence relations of default logic are *cautious reasoning* $\models_c$ and *brave reasoning* $\models_b$.

**Definition 3.5** *Let $\Delta = \langle D, W \rangle$ be a default theory and $\phi \in \mathcal{L}$ a formula. Then $\Delta \models_c \phi$ if and only if $\phi \in E$ for all extensions $E$ of $\Delta$, and $\Delta \models_b \phi$ if and only if $\phi \in E$ for some extension $E$ of $\Delta$.*

The following terminology is used in referring to default rules of certain syntactic forms. Defaults of the form $\alpha{:}\beta/\beta$, $\alpha{:}\beta \wedge \gamma/\gamma$, $\top{:}\beta_1, \ldots, \beta_n/\gamma$ are respectively *normal*, *seminormal*, and *prerequisite-free*. The symbol $\top$ denotes a valid formula. Prerequisite-free defaults are often written without prerequisites as $:\beta_1, \ldots, \beta_n/\gamma$. We shall sometimes denote sequences $\beta_1, \ldots, \beta_n$ of justifications by $\sigma, \sigma', \sigma_1$ and so on.

Not all seminormal default theories have extensions, but all *ordered* default theories do. In some cases, a decision problem for a class of default theories is intractable, but for the subclass in which the default theories are ordered, it is tractable (Kautz & Selman, 1991). In later sections we analyze the complexity of decision problems both with and without the orderedness condition.

**Definition 3.6 (Etherington, 1987)** *Let $\Delta = \langle D, W \rangle$ be a seminormal default theory. Without loss of generality assume that all formulae are in clausal form. The relations $\ll$ and $\lll$ are defined as follows.*

1. *If $\alpha \in W$, then $\alpha = \alpha_1 \vee \cdots \vee \alpha_n$ for some $n \geq 1$. For all $\alpha_i, \alpha_j \in \{\alpha_1, \ldots, \alpha_n\}$ such that $\alpha_i \neq \alpha_j$, let $\neg\alpha_i \lll \alpha_j$.*

2. *If $\delta \in D$, then $\delta = \alpha{:}\beta \wedge \gamma/\beta$. Let $A$, $B$ and $G$ be the sets of literals of the clausal forms of $\alpha$, $\beta$ and $\gamma$, respectively.*

   (a) *If $\alpha_i \in A$ and $\beta_j \in B$, let $\alpha_i \lll \beta_j$.*

(b) *If $\gamma_i \in G$ and $\beta_j \in B$, let $\neg\gamma_i \ll \beta_j$.*

(c) *Also, $\beta = \beta_1 \wedge \cdots \wedge \beta_m$ for some $m \geq 1$. For each $i \leq m$, $\beta_i = \beta_{i,1} \vee \cdots \vee \beta_{i,m_i}$ where $m_i \geq 1$. Thus if $\beta_{i,j}, \beta_{i,k} \in \{\beta_{1,1}, \cdots, \beta_{m,m_m}\}$ and $\beta_{i,j} \neq \beta_{i,k}$, let $\neg\beta_{i,j} \underline{\ll} \beta_{i,k}$.*

3. *The following transitivity relations hold for $\ll$ and $\underline{\ll}$.*

   (a) *If $\alpha \underline{\ll} \beta$ and $\beta \underline{\ll} \gamma$, then $\alpha \underline{\ll} \gamma$.*

   (b) *If $\alpha \ll \beta$ and $\beta \ll \gamma$, then $\alpha \ll \gamma$.*

   (c) *If $\alpha \ll \beta$ and $\beta \underline{\ll} \gamma$ or $\alpha \underline{\ll} \beta$ and $\beta \ll \gamma$, then $\alpha \ll \gamma$.*

*The default theory $\Delta$ is ordered if and only if $\alpha \ll \alpha$ for no $\alpha$.*

According to Theorem 1 in (Etherington, 1987), ordered default theories, like normal default theories and unlike seminormal default theories in general, have at least one extension.

## 4. Prioritized Default Logics by Brewka and by Baader and Hollunder

Priorities in default logic have been investigated by Baader and Hollunder (1995) and Brewka (1994). They view priorities as information that selects which defaults are applied next when constructing an extension. Priorities are strict partial orders, that is, transitive and asymmetric relations $\mathcal{P}$ on the defaults. If $\delta \mathcal{P} \delta'$, then the application of $\delta$ is more desirable than the application of $\delta'$, and the default $\delta$ is more significant or has a higher priority. Brewka, as well as Baader and Hollunder, give a definition of preferred extensions by modifying the semiconstructive definition of extensions (Theorem 3.2.) The construction of an extension starts from the set $W$ of objective facts, and proceeds in stages by repeatedly applying the highest priority applicable defaults. In this section we analyze the complexity of the associated decision problems.

**Definition 4.1** *A default $\alpha:\beta/\gamma$ is active in $E \subseteq \mathcal{L}$ if $E \models \alpha$ and $E \not\models \neg\beta$ and $E \not\models \gamma$.*

**Definition 4.2 (Baader and Hollunder, 1995)** *Let $\langle D, W \rangle$ be a default theory and $\mathcal{P}$ a strict partial order on $D$. Let $E \subseteq \mathcal{L}$ be a set of formulae. Define for all $i \geq 0$,*

$$
\begin{aligned}
E_0 &= W \\
E_{i+1} &= E_i \cup \left\{ \gamma \,\middle|\, \frac{\alpha:\beta}{\gamma} \in D, E_i \models \alpha, E \not\models \neg\beta, \text{ no } \delta \mathcal{P} \frac{\alpha:\beta}{\gamma} \text{ is active in } E_i \right\}.
\end{aligned}
$$

*Then $E$ is a $\mathcal{P}$-preferred$^{BH}$ extension of $\langle D, W \rangle$ if and only if $E = \bigcup_{i \geq 0} Cn(E_i)$.*

Brewka's definition is syntactically similar. We have expressed it in a way that highlights the differences to Baader and Hollunder's definition. The main difference is that the construction of a preferred extension proceeds by applying defaults in an order specified by some strict total order that extends the priorities. Furthermore, Brewka gives his definition for normal default theories only, and the consistency of justifications is tested against the sets $E_i$ instead of the set $E$.

**Definition 4.3 (Brewka, 1994)** *Let $\langle D, W \rangle$ be a normal default theory and $\mathcal{P}$ a strict partial order on $D$. Then $E \subseteq \mathcal{L}$ is a $\mathcal{P}$-preferred$^B$ extension of $\langle D, W \rangle$ if there is a strict total order $\mathcal{T}$ on $D$ such that $\mathcal{P} \subseteq \mathcal{T}$ and $E = \bigcup_{i \geq 0} Cn(E_i)$ where for all $i \geq 0$,*

$$E_0 = W$$
$$E_{i+1} = E_i \cup \left\{ \beta \,\middle|\, \frac{\alpha : \beta}{\beta} \in D, E_i \models \alpha, E_i \not\models \neg\beta, \ no \ \delta\mathcal{T}\frac{\alpha : \beta}{\beta} \ is \ active \ in \ E_i \right\}.$$

*We say that $E$ is generated by $\mathcal{T}$.*

The purpose of both of these definitions of preferred extensions is to introduce a mechanism for resolving conflicts between defaults on the basis of the priorities. Different extensions represent different ways of resolving the conflicts. The preferred extensions are a subset of all extensions of a default theory.

**Example 4.1** Let $\Delta = \langle D, W \rangle$ be a default theory and $\mathcal{P}$ a strict partial order on $D$, where

$$D = \left\{ \frac{priest : \neg like\text{-}rock\text{-}n\text{-}roll}{\neg like\text{-}rock\text{-}n\text{-}roll}, \frac{man : like\text{-}rock\text{-}n\text{-}roll}{like\text{-}rock\text{-}n\text{-}roll}, \frac{priest : man}{man} \right\},$$
$$W = \{priest\}, \ and$$
$$\mathcal{P} = \left\{ \left\langle \frac{priest : \neg like\text{-}rock\text{-}n\text{-}roll}{\neg like\text{-}rock\text{-}n\text{-}roll}, \frac{man : like\text{-}rock\text{-}n\text{-}roll}{like\text{-}rock\text{-}n\text{-}roll} \right\rangle \right\}.$$

The priorities $\mathcal{P}$ state that when reasoning about male priests, information specific to priests should override information concerning men in general. Of the extensions $E = Cn(\{priest, man, \neg like\text{-}rock\text{-}n\text{-}roll\})$ and $E' = Cn(\{priest, man, like\text{-}rock\text{-}n\text{-}roll\})$ only $E$ is $\mathcal{P}$-preferred$^B$ and $\mathcal{P}$-preferred$^{BH}$. In the Baader and Hollunder definition it is obtained as $E = \bigcup_{i \geq 0} Cn(E_i)$ where $E_0 = \{priest\}, E_1 = \{priest, man, \neg like\text{-}rock\text{-}n\text{-}roll\}$ and $E_i = E_1$ for all $i \geq 2$. In this extension, the conflict between the first two defaults is resolved in favor of the first one. $\square$

The definitions are not equivalent, as demonstrated by the following example given by Baader and Hollunder (1995). Baader and Hollunder also show that some extensions are preferred$^{BH}$ but not preferred$^B$.

**Example 4.2** Let $D = \{:a/a, :b/b, b{:}c/c, a{:}\neg c/\neg c\}$ and $\mathcal{P} = \{\langle b{:}c/c, a{:}\neg c/\neg c \rangle\}$. The extension $E = Cn(\{a, b, \neg c\})$ is $\mathcal{P}$-preferred$^B$ but not $\mathcal{P}$-preferred$^{BH}$. The extension $E$ cannot be obtained with the Baader and Hollunder definition. For $E' = Cn(\{a, b, c\})$ the Baader and Hollunder definition produces the sets $E'_0 = \emptyset, E'_1 = \{a, b\}, E'_2 = \{a, b, c\}$ and $E'_i = E'_2$ for all $i \geq 3$, and the unique $\mathcal{P}$-preferred$^{BH}$ extension is $E' = \bigcup_{i \geq 0} E'_i$. In Brewka's definition we can bypass the higher priority default $b{:}c/c$ by using a total ordering where $:b/b$ follows $:a/a$ and $a{:}\neg c/\neg c$. This way we obtain the sets $E_0 = \emptyset$, $E_1 = \{a\}$, $E_2 = \{a, \neg c\}$, $E_3 = \{a, \neg c, b\}$ and $E_i = E_3$ for all $i \geq 4$. $\square$

Consequence relations that correspond to cautious reasoning in Reiter's default logic are defined as follows.

**Definition 4.4** *The consequence relations $\models^B$ and $\models^{BH}$ are defined by $\Delta \models^B_{\mathcal{P}} \phi$ if and only if the formula $\phi$ is in all $\mathcal{P}$-preferred$^B$ extensions of $\Delta$, and $\Delta \models^{BH}_{\mathcal{P}} \phi$ if and only if the formula $\phi$ is in all $\mathcal{P}$-preferred$^{BH}$ extensions of $\Delta$.*

**Lemma 4.5** *Let $\langle D, W \rangle$ be a normal default theory and $\mathcal{T}$ a strict total order on $D$. Then for all $E \subseteq \mathcal{L}$, $E$ is a $\mathcal{T}$-preferred$^B$ extension of $\langle D, W \rangle$ if and only if $E$ is a $\mathcal{T}$-preferred$^{BH}$ extension of $\langle D, W \rangle$.*

*Proof:* If $W$ is inconsistent, then the inconsistent extension $\mathcal{L}$ is the unique $\mathcal{T}$-preferred$^B$ and $\mathcal{T}$-preferred$^{BH}$ extension of $\Delta$. So assume $W$ is consistent. ($\Rightarrow$) Assume that $E$ is a $\mathcal{T}$-preferred$^B$ extension of $\langle D, W \rangle$. By definition $E = \bigcup_{i \geq 0} Cn(E_i)$, where $E_i$ are as in Definition 4.3. Let $E'_i, i \geq 0$ be the sets $E_i$ in Definition 4.2. We show that $E_i = E'_i$ for all $i \geq 0$, and hence $E = \bigcup_{i \geq 0} Cn(E'_i)$ is a $\mathcal{T}$-preferred$^{BH}$ extension of $\Delta$. *Induction hypothesis*: $E_i = E'_i$. *Base case $i = 0$*: Immediate. *Inductive case $i \geq 1$*: Assume that for $\alpha{:}\beta/\beta \in D$, $E_{i-1} \models \alpha$, $E_{i-1} \not\models \neg\beta$ and no $\delta \mathcal{T} \alpha{:}\beta/\beta$ is active in $E_{i-1}$. Hence $\beta \in E_i$. By the induction hypothesis $E'_{i-1} \models \alpha$ and no $\delta \mathcal{T} \alpha{:}\beta/\beta$ is active in $E'_{i-1}$. To show $\beta \in E'_i$ it remains to show that $E \not\models \neg\beta$. Because $E$ is consistent and $E_i \subseteq E$ and $\beta \in E_i$, $E \not\models \neg\beta$. Hence $\beta \in E'_i$ and $E_i \subseteq E'_i$. Proof of $E'_i \subseteq E_i$ is similar. It suffices to point out that $E \not\models \neg\beta$ implies $E_{i-1} \not\models \neg\beta$ simply because $E_{i-1} \subseteq E$.

($\Leftarrow$) Let $E$ be a $\mathcal{T}$-preferred$^{BH}$ extension of $\langle D, W \rangle$. Hence $E = \bigcup_{i \geq 0} Cn(E'_i)$, where $E'_i, i \geq 0$ are the sets in Definition 4.2. Let $E_i, i \geq 0$ be the sets in Definition 4.3. We show by induction that $E = \bigcup_{i \geq 0} E_i$. *Induction hypothesis*: $E_i = E'_i$. *Base case $i = 0$*: Immediate. *Inductive case $i \geq 1$*: Assume that for $\alpha{:}\beta/\beta \in D$, $E'_{i-1} \models \alpha$, $E \not\models \neg\beta$ and no $\delta \mathcal{T} \alpha{:}\beta/\beta$ is active in $E'_{i-1}$. Hence $\beta \in E'_i$. By the induction hypothesis $E_{i-1} \models \alpha$ and no $\delta \mathcal{T} \alpha{:}\beta/\beta$ is active in $E_{i-1}$. To show $\beta \in E_i$ it remains to show that $E_{i-1} \not\models \neg\beta$. Because $E'_{i-1} \subseteq E$ and $E \not\models \neg\beta$, $E'_{i-1} \not\models \neg\beta$. By the induction hypothesis $E_{i-1} = E'_{i-1}$. Hence $E_{i-1} \not\models \neg\beta$. Therefore $E'_i \subseteq E_i$. Proof of $E_i \subseteq E'_i$ proceeds similarly. Assume that for $\alpha{:}\beta/\beta \in D$, $E_{i-1} \models \alpha$, $E_{i-1} \not\models \neg\beta$ and no $\delta \mathcal{T} \alpha{:}\beta/\beta$ is active in $E_{i-1}$. By the induction hypothesis $E'_{i-1} \models \alpha$ and no $\delta \mathcal{T} \alpha{:}\beta/\beta$ is active in $E'_{i-1}$. It remains to show that $E \not\models \neg\beta$. So assume $E \models \neg\beta$. Because $\alpha{:}\beta/\beta$ is the $\mathcal{T}$-least active default and $E \models \neg\beta$, $E'_i = E'_{i-1}$, and further $E'_j = E'_{i-1}$ for all $j \geq i$. Therefore $E = Cn(E'_{i-1})$. This leads to a contradiction with the assumption $E \models \neg\beta$ and the fact $E'_{i-1} \not\models \neg\beta$ obtained with the induction hypothesis. Therefore the assumption is false, and $E \not\models \neg\beta$. Therefore $\beta \in E'_i$ and $E_i \subseteq E'_i$. $\square$

## 4.1 Complexity in the General Case

With no restrictions on the form of defaults, both the Brewka and the Baader and Hollunder prioritized default logics are complete for the second level of the polynomial hierarchy. This means that there are polynomial time translations between cautious reasoning in Reiter's default logic and the prioritized versions of cautious reasoning in both of these logics.

**Theorem 4.6** *Testing $\Delta \models^B_{\mathcal{P}} \phi$ for normal default theories $\Delta$, strict partial orders $\mathcal{P}$ and formulae $\phi$ is $\Pi^p_2$-complete.*

*Proof:* The $\Pi^p_2$-hardness is because with an empty priority relation Brewka's logic coincides with Reiter's default logic (Proposition 6 in (Brewka, 1994)), which is $\Pi^p_2$-hard even with the restriction to normal defaults (Gottlob, 1992). We show that the complement of the problem is in $\Sigma_2^p$, which directly implies that the problem is in $\Pi_2^p$. The complementary problem is the existence of a $\mathcal{P}$-preferred$^B$ extension $E$ such that $\phi \notin E$. The guessing of a strict total order $\mathcal{T}$ that generates an extension $E$ such that $\phi \notin E$ can be done by a nondeterministic Turing machine in a polynomial number of steps: guess $\mathcal{T}$, and guess the conclusions of generating defaults of $E$. The verification that $E$ fulfills Brewka's definition takes a polynomial number of steps with an NP oracle for propositional satisfiability. □

| | class of default theories | form of defaults |
|---|---|---|
| 1 | disjunction-free | $\frac{a_1 \wedge \cdots \wedge a_i : b_1 \wedge \cdots \wedge b_n \wedge c_1 \wedge \cdots \wedge c_m}{b_1 \wedge \cdots \wedge b_n}$ |
| 2 | unary | $\frac{p:q}{q} \quad \frac{p:q \wedge \neg r}{q} \quad \frac{p:\neg q}{\neg q}$ |
| 3 | disjunction-free ordered | $\frac{a_1 \wedge \cdots \wedge a_i : b_1 \wedge \cdots \wedge b_n \wedge c_1 \wedge \cdots \wedge c_m}{b_1 \wedge \cdots \wedge b_n}$ |
| 4 | ordered unary | $\frac{p:q}{q} \quad \frac{p:q \wedge \neg r}{q} \quad \frac{p:\neg q}{\neg q}$ |
| 5 | disjunction-free normal | $\frac{a_1 \wedge \cdots \wedge a_i : b_1 \wedge \cdots \wedge b_n}{b_1 \wedge \cdots \wedge b_n}$ |
| 6 | Horn | $\frac{p_1 \wedge \cdots \wedge p_n : q}{q} \quad \frac{p_1 \wedge \cdots \wedge p_n : \neg q}{\neg q}$ |
| 7 | normal unary | $\frac{p:q}{q} \quad \frac{p:\neg q}{\neg q}$ |
| 8 | prerequisite-free | $\frac{:b_1 \wedge \cdots \wedge b_n \wedge c_1 \wedge \cdots \wedge c_m}{b_1 \wedge \cdots \wedge b_n}$ |
| 9 | prerequisite-free ordered | $\frac{:b_1 \wedge \cdots \wedge b_n \wedge c_1 \wedge \cdots \wedge c_m}{b_1 \wedge \cdots \wedge b_n}$ |
| 10 | prerequisite-free unary | $\frac{:q}{q} \quad \frac{:q \wedge \neg r}{q} \quad \frac{:\neg q}{\neg q}$ |
| 11 | prerequisite-free ordered unary | $\frac{:q}{q} \quad \frac{:q \wedge \neg r}{q} \quad \frac{:\neg q}{\neg q}$ |
| 12 | prerequisite-free normal | $\frac{:b_1 \wedge \cdots \wedge b_n}{b_1 \wedge \cdots \wedge b_n}$ |
| 13 | prerequisite-free normal unary | $\frac{:q}{q} \quad \frac{:\neg q}{\neg q}$ |
| 14 | prerequisite-free positive normal unary | $\frac{:q}{q}$ |

Table 1: Form of defaults in a number of classes of default theories

**Theorem 4.7** *Testing* $\Delta \models_{\mathcal{P}}^{BH} \phi$ *for default theories* $\Delta$, *strict partial orders* $\mathcal{P}$ *and formulae* $\phi$ *is* $\Pi_2^p$-complete.

*Proof:* Like the proof of the previous theorem except that no strict total orders need to be guessed. □

## 4.2 Complexity in Syntactically Restricted Cases

In this section we investigate the complexity of the prioritized default logics in syntactically restricted cases. This subject is related to the research on the boundary between tractability and intractability of default logic without priorities by Kautz and Selman (1991) and Stillman (1990). First we analyze the case in which the priorities are strict total orders. Then we complete the complexity analysis by extending the results to arbitrary strict partial orders.

Kautz and Selman analyze the complexity of determining the existence of extensions, cautious reasoning, and brave reasoning for default theories with defaults of the form listed in Table 1 (excluding the prerequisite-free classes) and with objective parts that are sets of literals. In Table 1, the letters $a, b, c$ – possibly subscripted – denote literals, and the letters $p, q, r$ denote propositional variables. Stillman considers only the problem of brave reasoning, but considers a wider range of classes. He analyzes default theories the objective parts of which are sets of Horn clauses or sets of 2-literal clauses, and also the case where defaults are prerequisite-free.[1] The most general tractable classes in default logic without priorities are the following.

- If the objective part $W$ may contain Horn clauses, none of the restrictions on the form of default rules is sufficient for achieving tractability in brave reasoning (Stillman, 1990). Cautious reasoning has not been investigated.

- If $W$ consists of 2-literal clauses, brave reasoning for the prerequisite-free normal class is tractable (Stillman, 1990). Cautious reasoning has not been investigated.

- If $W$ consists of literals, brave reasoning for the prerequisite-free normal class and the Horn class are tractable (Kautz & Selman, 1991; Stillman, 1990). Cautious reasoning is tractable for the normal unary class (Kautz & Selman, 1991). Cautious reasoning for the prerequisite-free classes has not been investigated.

We present a similar analysis for the same classes of prioritized default logics starting with the case where priorities totally order the defaults. We present for each class either a polynomial time decision procedure, or an NP-hardness or a co-NP-hardness result.

### 4.2.1 Summary

Because of the constructive nature of the definitions of preferred extensions in the prioritized default logics by Brewka and by Baader and Hollunder, constructing the unique preferred extensions of default theories with total priorities is tractable whenever the logical consequence tests in propositional logic are tractable (Theorem 4.8.) This is because there is always a unique default that is applied next, and this default cannot be defeated by any default applied later. So with total priorities, these prioritized default logics are computationally much easier than Reiter's default logic.

For unrestricted priorities, Table 2 summarizes the complexity of $\models^B$ and $\models^{BH}$ with various syntactic restrictions. As Brewka does not consider non-normal default theories, complexity results on seminormal classes concern the consequence relation $\models^{BH}$ only. In cases where Reiter's default logic is tractable and a prioritized default logic is not, or vice versa, the complexity characterization is set in boldface.

Table 3 gives references to the theorems and corollaries where the results are stated. We use the notation $n \subseteq$ to indicate that the intractability of the class is directly implied by the intractability of the subclass $n$ in the same column, and the notation $\subseteq n$ to indicate that the tractability is implied by the tractability of the superclass $n$ in the same column. When the complexity is directly implied by the complexity of an unprioritized class of default theories investigated by Kautz and Selman (1991) we indicate this by K&S.

---

1. Stillman's definition of prerequisite-free unary and prerequisite-free ordered unary classes includes only default rules of the forms $:\neg q/\neg q$ and $:p \wedge \neg q/p$. We, however, consider also default rules of the form $:p/p$ in order to have a closer correspondence with the Kautz and Selman definition of unary classes. This change does not sacrifice generality, as the default $:p/p$ works like $:p \wedge \neg q/p$ whenever $q$ does not occur elsewhere in the default theory.

| | class of default theories | complexity when clauses in W are | | |
|---|---|---|---|---|
| | | Horn | 2-literal | 1-literal |
| 1 | disjunction-free | co-NP-hard | co-NP-hard | co-NP-hard |
| 2 | unary | co-NP-hard | co-NP-hard | co-NP-hard |
| 3 | disjunction-free ordered | co-NP-hard | co-NP-hard | co-NP-hard |
| 4 | ordered unary | co-NP-hard | co-NP-hard | co-NP-hard |
| 5 | disjunction-free normal | co-NP-hard | co-NP-hard | co-NP-hard |
| 6 | Horn | co-NP-hard | co-NP-hard | co-NP-hard |
| 7 | normal unary | co-NP-hard | co-NP-hard | **co-NP-hard** |
| 8 | prerequisite-free | co-NP-hard | co-NP-hard | co-NP-hard |
| 9 | prerequisite-free ordered | co-NP-hard | co-NP-hard | co-NP-hard |
| 10 | prerequisite-free unary | co-NP-hard | co-NP-hard | co-NP-hard |
| 11 | prerequisite-free ordered unary | co-NP-hard | co-NP-hard | co-NP-hard |
| 12 | prerequisite-free normal | co-NP-hard | co-NP-hard | co-NP-hard |
| 13 | prerequisite-free normal unary | co-NP-hard | co-NP-hard | PTIME |
| 14 | prerequisite-free positive normal unary | co-NP-hard | co-NP-hard | PTIME |

Table 2: Complexity of the consequence relations $\models^B$ and $\models^{BH}$

| | class of default theories | reference | | |
|---|---|---|---|---|
| | | Horn | 2-literal | 1-literal |
| 1 | disjunction-free | K&S | K&S | K&S |
| 2 | unary | K&S | K&S | K&S |
| 3 | disjunction-free ordered | K&S | K&S | K&S |
| 4 | ordered unary | K&S | K&S | K&S |
| 5 | disjunction-free normal | K&S | K&S | K&S |
| 6 | Horn | K&S | K&S | K&S |
| 7 | normal unary | 14 $\subseteq$ | 14 $\subseteq$ | T4.15 |
| 8 | prerequisite-free | 14 $\subseteq$ | 14 $\subseteq$ | 12 $\subseteq$ |
| 9 | prerequisite-free ordered | 14 $\subseteq$ | 14 $\subseteq$ | 12 $\subseteq$ |
| 10 | prerequisite-free unary | 14 $\subseteq$ | 14 $\subseteq$ | 11 $\subseteq$ |
| 11 | prerequisite-free ordered unary | 14 $\subseteq$ | 14 $\subseteq$ | T4.14 |
| 12 | prerequisite-free normal | 14 $\subseteq$ | 14 $\subseteq$ | T4.13 |
| 13 | prerequisite-free normal unary | 14 $\subseteq$ | 14 $\subseteq$ | T4.11 |
| 14 | prerequisite-free positive normal unary | T4.12 | T4.12 | $\subseteq$ 13 |

Table 3: References to theorems on the complexity of $\models^B$ and $\models^{BH}$

When looking at Table 2 it is slightly surprising that the favorable computational properties in the totally ordered case are in no way reflected in the complexity of reasoning with arbitrary priorities. With arbitrary priorities, for almost all classes of default theories in the Kautz-Selman-Stillman hierarchy the complexity is the same as the complexity of Reiter's default logic and the lexicographic prioritized default logic that is discussed in Section 5.

#### 4.2.2 Tractable Classes

The following theorem shows that with the restriction to priorities that totally order the defaults, both the Baader and Hollunder and the Brewka definition of preferred extensions yield an efficient decision procedure for the respective prioritized default logic.

**Theorem 4.8** *Let $\langle D, W \rangle$ be a finite default theory and $\mathcal{T}$ a strict total order on $D$. The unique $\mathcal{T}$-preferred$^{BH}$ extension (if preferred$^{BH}$ extensions exist) of $\langle D, W \rangle$ can be computed in polynomial time if all the membership tests in $Cn(E_i)$ in the definition of $E$ are for a polynomial time subset of the propositional logic.*

*Proof:* Unlike the definition of preferred$^{B}$ extensions, the definition of preferred$^{BH}$ extensions does not directly yield a polynomial time decision procedure for total priorities and tractable classical reasoning. This is because the test that the justifications of defaults to be applied do not belong to the extension being constructed cannot be performed before the extension is fully known. However, if a simple test for justifications of defaults is added, the algorithm that works with preferred$^{B}$ extensions works also with preferred$^{BH}$ extensions. Compute sets $E_i'$ as follows.

$$E_0' = W$$
$$E_{i+1}' = E_i' \cup \left\{ \gamma \,\middle|\, \frac{\alpha : \beta}{\gamma} \in D, E_i' \models \alpha, E_i' \not\models \neg\beta, \text{ no } \delta\mathcal{T}\frac{\alpha : \beta}{\gamma} \text{ is active in } E_i' \right\}$$

If $E_i' \models \neg\beta$ for some $\alpha{:}\beta/\gamma \in D$ such that for some $j < i$, $E_j' \models \alpha$, $E_j' \not\models \neg\beta$ and no $\delta\mathcal{T}\alpha{:}\beta/\gamma$ is active in $E_j'$, then the algorithm returns *false*. If $E_i' = E_{i+1}'$ for some $i \geq 0$, then return $Cn(E_i') = \bigcup_{j \in \{0,\dots,i\}} Cn(E_j')$ and define $E_j' = E_i'$ for all $j > i$.

Let $E$ be a $\mathcal{T}$-preferred$^{BH}$ extension of $\langle D, W \rangle$. Hence $E = \bigcup_{i \geq 0} Cn(E_i)$, where $E_i, i \geq 0$ are the sets in Definition 4.2. We show by induction that $E = \bigcup_{i \geq 0} E_i'$ and that the value *false* is not returned. *Induction hypothesis*: $E_i' = E_i$. *Base case $i = 0$*: Immediate. *Inductive case $i \geq 1$*: We first show that $E_i \subseteq E_i'$. Assume that for $\alpha{:}\beta/\gamma \in D$, $E_{i-1} \models \alpha$, $E \not\models \neg\beta$ and no $\delta\mathcal{T}\alpha{:}\beta/\gamma$ is active in $E_{i-1}$. Hence $\gamma \in E_i$. By the induction hypothesis $E_{i-1}' \models \alpha$ and no $\delta\mathcal{T}\alpha{:}\beta/\gamma$ is active in $E_{i-1}'$. To show $\gamma \in E_i'$ it remains to show that $E_{i-1}' \not\models \neg\beta$. Because $E_{i-1} \subseteq E$ and $E \not\models \neg\beta$, $E_{i-1} \not\models \neg\beta$. By the induction hypothesis $E_{i-1}' = E_{i-1}$. Hence $E_{i-1}' \not\models \neg\beta$. Therefore $\gamma \in E_i'$ and $E_i \subseteq E_i'$. Proof of $E_i' \subseteq E_i$ proceeds similarly. Assume that for $\alpha{:}\beta/\gamma \in D$, $E_{i-1}' \models \alpha$, $E_{i-1}' \not\models \neg\beta$ and no $\delta\mathcal{T}\alpha{:}\beta/\gamma$ is active in $E_{i-1}'$. By the induction hypothesis $E_{i-1} \models \alpha$ and no $\delta\mathcal{T}\alpha{:}\beta/\gamma$ is active in $E_{i-1}$. It remains to show that $E \not\models \neg\beta$. So assume $E \models \neg\beta$. Because $\alpha{:}\beta/\gamma$ is the $\mathcal{T}$-least default and $E \models \neg\beta$, $E_i = E_{i-1}$, and further $E_j = E_{i-1}$ for all $j \geq i$. Therefore $E = Cn(E_{i-1})$. This leads to a contradiction with the assumption $E \models \neg\beta$ and the fact $E_{i-1} \not\models \neg\beta$ obtained with the induction hypothesis. Therefore the assumption is false, and $E \not\models \neg\beta$. Therefore $\gamma \in E_i$ and $E_i' \subseteq E_i$. That the algorithm does not return *false* goes similarly. Assume that for some $\alpha{:}\beta/\gamma \in D$, $E_i' \models \neg\beta$ and for some $j < i$, $E_j' \models \alpha$ and no $\delta\mathcal{T}\alpha{:}\beta/\gamma$ is active in $E_j'$. We have to show that $E_j' \not\models \neg\beta$. This is implied by the fact $E \not\models \neg\beta$ shown above because $E_j' \subseteq E$.

Assume that the algorithm yields $E = \bigcup_{i \geq 0} Cn(E_i')$. We claim that $E$ is a $\mathcal{P}$-preferred$^{BH}$ extension of $\langle D, W \rangle$. Because the algorithm did not return *false* and $E = \bigcup_{i \geq 0} Cn(E_i')$,

```
PROCEDURE decide(l, D, W)
   IF W ⊨ l THEN RETURN true;
   IF l̄ ∈ W THEN RETURN false;
   IF :l/l ∉ D THEN RETURN false;
   IF :l̄/l̄ ∉ D THEN RETURN true;
   IF :l/l𝒫:l̄/l̄ THEN RETURN true;
   RETURN false
END
```

Figure 1: A decision procedure for prioritized prerequisite-free normal unary theories

$E \not\models \neg\beta$ for all $\alpha{:}\beta/\gamma \in D$ such that for some $j$, $E'_j \models \alpha$, $E'_j \not\models \neg\beta$ and no $\delta\mathcal{T}\alpha{:}\beta/\gamma$ is active in $E'_j$. It is straightforward to show that $E'_i = E_i$ for all $i \geq 0$, where $E_i$ are the sets in Definition 4.2. Therefore $E$ is a $\mathcal{T}$-preferred$^{BH}$ extension of $\langle D, W \rangle$.  □

This theorem indicates that with a restriction to total priorities $\mathcal{T}$, testing the membership of a literal in all $\mathcal{T}$-preferred$^{BH}$ extensions of default theories in all the classes of the Kautz and Selman and Stillman hierarchy is in P.

**Corollary 4.9** *Testing $\langle D, W \rangle \models^{BH}_{\mathcal{P}} \phi$ for disjunction-free default theories $\langle D, W \rangle$, where $W$ is a set of Horn clauses or 2-literal clauses, $\mathcal{P}$ is a strict total order on $D$, and $\phi$ is a literal, can be done in polynomial time.*

**Corollary 4.10** *Testing $\langle D, W \rangle \models^{B}_{\mathcal{P}} \phi$ for disjunction-free normal default theories $\langle D, W \rangle$, where $W$ is a set of Horn clauses or 2-literal clauses, $\mathcal{P}$ is a strict total order on $D$, and $\phi$ is a literal, can be done in polynomial time.*

*Proof:* Brewka gives a definition of preferred extensions for normal default theories only. By Lemma 4.5 the preferred extensions in this case coincide with the Baader and Hollunder preferred extensions whenever the priorities are a strict total order.  □

For unrestricted priorities the class of prerequisite-free normal unary default theories is tractable. The remaining classes in the hierarchy are intractable.

**Theorem 4.11** *For literals $l$ and prerequisite-free normal unary theories $\langle D, W \rangle$ where $W$ is a set of literals, testing $\langle D, W \rangle \models^{BH}_{\mathcal{P}} l$ and $\langle D, W \rangle \models^{B}_{\mathcal{P}} l$ can be done in polynomial time.*

*Proof:* The algorithm in Figure 1 tests $\langle D, W \rangle \models^{B}_{\mathcal{P}} l$ and $\langle D, W \rangle \models^{BH}_{\mathcal{P}} l$. The correctness of the algorithm for $\models^{BH}_{\mathcal{P}}$ is as follows. For $\models^{B}_{\mathcal{P}}$ the proof is similar. We analyze the *if*-statements in sequence. In each case we may use the negations of the assumptions of the previous cases.

1. Assume that $W \models l$. Now $l$ is in all $\mathcal{P}$-preferred$^{BH}$ extensions $E$ of $\Delta$ because by definition $W \subseteq E$. Hence it is correct to return *true*. 2. Assume that $\bar{l} \in W$. Because $W$ is consistent (as $W \not\models l$), all extensions are consistent, and no extension contains $l$. Hence it is correct to return *false*. 3. Assume that $:l/l \notin D$. Because $l \notin W$ and all extensions are consistent, no extension contains $l$. Hence it is correct to return *false*. 4. Assume that $:\bar{l}/\bar{l} \notin D$. Because $\bar{l} \notin W$, no extension contains $\bar{l}$, and hence $:l/l$ is applied in all extensions, and $l$ is in all extensions. Hence it is correct to return *true*. 5. Assume that $:l/l\mathcal{P}:\bar{l}/\bar{l}$. Assume that there is a preferred$^{BH}$ extension $E$ such that $\bar{l} \in E$. Now $\bar{l} \in E_i \backslash E_{i-1}$ for some $i$. Hence $l \notin E_{i-1}$ and $:l/l$ is not active in $E_{i-1}$. Hence $\neg l \in E_{i-1}$, which however contradicts the fact that $\bar{l} \notin E_{i-1}$. Hence $\bar{l}$ in no $\mathcal{P}$-preferred$^{BH}$ extension of $\Delta$, and $\Delta \models_{\mathcal{P}}^{BH} l$. Hence it is correct to return *true*. 6. In the remaining case not $:l/l\mathcal{P}:\bar{l}/\bar{l}$. If $:\bar{l}/\bar{l}\mathcal{P}:l/l$, then by an argument similar to the previous case all $\mathcal{P}$-preferred$^{BH}$ extensions – by assumption there is at least one – contain $\bar{l}$, and hence it is correct to return *false*. If neither $:\bar{l}/\bar{l}\mathcal{P}:l/l$ nor $:l/l\mathcal{P}:\bar{l}/\bar{l}$, then by symmetry there is an extension not containing $l$, and hence it is correct to return *false*.

Therefore the algorithm returns true if and only if $l$ is in all $\mathcal{P}$-preferred$^{BH}$ extensions of $\Delta$. The algorithm obviously runs in polynomial time. □

### 4.2.3 INTRACTABLE CLASSES

Intractability of all remaining classes except the normal unary class is directly implied by the intractability of the same classes in Reiter's default logic, as shown by Kautz and Selman (1991) and Theorems 4.12, 4.13 and 4.14.

Stillman (1990) analyzes the complexity of prerequisite-free default theories, and claims that brave reasoning for the prerequisite-free normal class with 2-literal clauses is solvable in polynomial time. However, he does not analyze the complexity of cautious reasoning. It turns out that even with the restriction to prerequisite-free normal defaults with propositional variables in justifications and conclusions, reasoning is intractable.

**Theorem 4.12** *Testing* $\langle D, W \rangle \models_c l$ *for literals* $l$ *and prerequisite-free positive normal unary default theories with objective parts that consist of 2-literal Horn clauses, is co-NP-hard.*

*Proof:* We give a many-one reduction from propositional satisfiability to the complement of the problem. Let $C$ be a set of clauses and $P$ the set of propositional variables that occur in $C$. Let $N$ be an injective function that maps each clause $c \in C$ to a propositional variable $n = N(c)$ such that $n \notin P$. Let

$$D = \left\{ \frac{:p'}{p'} \,\middle|\, p \in P \right\} \cup \left\{ \frac{:p''}{p''} \,\middle|\, p \in P \right\} \cup \left\{ \frac{:n}{n} \,\middle|\, c \in C, n = N(c) \right\}, \text{and}$$

$$W = \{p' \rightarrow p | p \in P\} \cup \{p'' \rightarrow \neg p | p \in P\}$$
$$\cup \{p \rightarrow \neg n | p \in c \in C, n = N(c)\} \cup \{\neg p \rightarrow \neg n | \neg p \in c \in C, n = N(c)\}$$
$$\cup \{n \rightarrow false | c \in C, n = N(c)\}.$$

We claim that $\langle D, W \rangle \not\models_c false$ if and only if $C$ is satisfiable.

Assume that $C$ is satisfiable; that is, there is a model $M$ such that $M \models C$. We show that there is an extension of $\langle D, W \rangle$ that does not contain *false*. Let $E = Cn(\{p'|p \in P, M \models p\} \cup \{p''|p \in P, M \not\models p\} \cup W)$. To show that $E$ is an extension of $\langle D, W \rangle$, it suffices to show that $E$ is consistent and for all $:\phi/\phi \in D$, $\neg \phi \notin E$ if and only if $\phi \notin E$. Let $M'$ be a model such that for all $p \in P$, $M' \models p$ iff $M \models p$, $M' \models p'$ iff $M \models p$, and $M' \models p''$ iff $M \not\models p$, and $M' \not\models n$ for all $n$ such that $n = N(c)$ for some $c \in C$. It is straightforward to show that $M' \models E$, and $E$ is therefore consistent. Take any $:\phi/\phi \in D$. Assume that

436

$\neg\phi \in E$. Because $E$ is consistent, $\phi \notin E$. Assume that $\phi \notin E$. If $\phi = p'$, then by definition $p'' \in E$, and as $\{p'' \to \neg p, p' \to p\} \subseteq E$, $\neg p' \in E$. Similarly for $\phi = p''$. If $\phi = n$ such that $n = N(c)$ for some $c \in C$, then because $M \models c$, there is disjunct $l \in \{p, \neg p\}$ of $c$ such that $M \models l$, and hence by definition $l \in E$, and as $l \to \neg n \in E$, $\neg n \in E$. Therefore $E$ is an extension of $\langle D, W \rangle$.

Assume that $\langle D, W \rangle \not\models_c false$; that is, there is an extension $E$ of $\langle D, W \rangle$ such that $false \notin E$. Let $M$ be a model such that for all $p \in P$, $M \models p$ iff $p \in E$. We show that $M \models C$, and hence $C$ is satisfiable. Because $false \notin E$ and $n \to false \in E$ for all $n$ such that $n = N(c)$ for some $c \in C$, $n \in E$ for no $n$ such that $n = N(c)$ for some $c \in C$. Because $:n/n \in D$ for all such $n$, $\neg n \in E$ for all such $n$, which means that for every clause in $C$, one of its disjuncts is in $E$. By definition this disjunct is true in $M$. Hence every clause in $C$ is true in $M$. □

An alternative way of obtaining the intractability of cautious reasoning for all classes with a Horn objective part is to apply Theorem 8.2 in (Kautz & Selman, 1991) – that reduces brave reasoning to cautious reasoning by adding a default $:l/l$ – and the intractability result for brave reasoning in Horn classes (Stillman, 1990). This is not applicable to the 2-literal case because in it brave reasoning is tractable.

**Theorem 4.13** *Testing $\langle D, \emptyset \rangle \models_c l$ for literals $l$ and sets $D$ of prerequisite-free normal defaults with conclusions that are conjunctions of literals, is co-NP-hard.*

*Proof:* Let $C$ be a set of clauses. We show that there is a default theory $\langle D, \emptyset \rangle$ such that $\langle D, \emptyset \rangle \not\models_c false$ if and only if $C$ is satisfiable. Let $P$ be the set of propositional variables that occur in $C$. Let $N$ be an injective function that maps each clause $c \in C$ to a propositional variable $n = N(c)$ such that $n \notin P$. Define the set of defaults $D$ as follows.

$$D = \left\{ \frac{:l \wedge n}{l \wedge n} \,\middle|\, c \in C, l \in c, n = N(c) \right\} \cup \left\{ \frac{:\neg n \wedge false}{\neg n \wedge false} \,\middle|\, c \in C, n = N(c) \right\}$$

Assume that $C$ is satisfiable. Let $M$ be a model such that $M \models C$. Let

$$E' = Cn(\{l \wedge n | c \in C, l \in c, n = N(c), M \models l\}).$$

Obviously $E'$ is consistent and it does not contain $false$.[2] Clearly $E'$ is an extension of $\langle\{:l \wedge n/l \wedge n \in D | c \in C, l \in c, n = N(c), l \wedge n \in E'\}, \emptyset\rangle$. By Theorem 3.2 in (Reiter, 1980) there is an extension $E$ of $\langle D, \emptyset \rangle$ such that $E' \subseteq E$. Because $n \in E' \subseteq E$ for all $n = N(c)$ with $c \in C$, $false \notin E$.

Assume that there is an extension $E$ of $\langle D, \emptyset \rangle$ such that $false \notin E$. Let $M$ be a model such that for all propositional variables $p$, $M \models p$ if and only if $p \in E$. Because $false \notin E$, $:\neg n \wedge false/\neg n \wedge false \in GD(E, \langle D, \emptyset \rangle)$ for no $n = N(c), c \in C$. Take any $c \in C$. Now $n = N(c) \in E$ and hence $:l \wedge n/l \wedge n \in GD(E, \langle D, \emptyset \rangle)$ for some $l \in c$. Hence $l \wedge n \in E$ and $M \models c$. Because this holds for all $c \in C$, finally $M \models C$. □

---

2. $E'$ is not necessarily an extension of $\langle D, \emptyset \rangle$. Consider the satisfiable formula $a \vee b$ the clausal form of which is $C = \{\{a, b\}\}$, and the model $M$ that assigns *true* to $a$ and *false* to $b$. Now $E' = Cn(\{a \wedge n\})$. However, there is the set $E = Cn(\{a \wedge n, b \wedge n\})$ that extends $E'$ and is an extension of $\langle D, \emptyset \rangle$.

**Theorem 4.14** *Testing $\langle D, \emptyset \rangle \models_c l$ for literals $l$ and sets $D$ of prerequisite-free ordered unary defaults is co-NP-hard.*

*Proof:* Proof is by many-one reduction from propositional satisfiability to testing $\langle D, \emptyset \rangle \not\models_c l$. Let $C$ be any set of clauses, and let $P$ be the set of propositional variables occurring in $C$. Let $N$ be an injective function that maps each clause $c \in C$ to a propositional variable $n = N(c)$ such that $n \notin P$. Let

$$
\begin{aligned}
D \;=\; & \left\{ \frac{:p}{p} \,\middle|\, p \in P \right\} \cup \left\{ \frac{:\neg p}{\neg p} \,\middle|\, p \in P \right\} \cup \left\{ \frac{:p' \wedge \neg p}{p'} \,\middle|\, p \in P \right\} \\
& \cup \left\{ \frac{:n \wedge \neg p}{n} \,\middle|\, c \in C, n = N(c), \neg p \in c \right\} \\
& \cup \left\{ \frac{:n \wedge \neg p'}{n} \,\middle|\, c \in C, n = N(c), p \in c \right\} \cup \left\{ \frac{:false \wedge \neg n}{false} \,\middle|\, c \in C, n = N(c) \right\}.
\end{aligned}
$$

The orderedness condition is fulfilled because the relation $\ll \,\subseteq\, (P \times \{N(c)|c \in C\}) \cup (\{p'|p \in P\} \times \{N(c)|c \in C\}) \cup (\{N(c)|c \in C\} \times \{false\}) \cup (P \times \{p'|p \in P\})$ is irreflexive.

We claim that $C$ is satisfiable if and only if $\langle D, \emptyset \rangle \not\models_c false$. Assume that $C$ is satisfiable; that is, there is a model $M$ such that $M \models C$. We construct an extension $E$ of $\langle D, \emptyset \rangle$ such that $false \notin E$. Let

$$
\begin{aligned}
A \;=\; & \left\{ \frac{:p}{p} \,\middle|\, p \in P, M \models p \right\} \cup \left\{ \frac{:\neg p}{\neg p} \,\middle|\, p \in P, M \not\models p \right\} \\
& \cup \left\{ \frac{:p' \wedge \neg p}{p'} \,\middle|\, p \in P, M \not\models p \right\} \\
& \cup \left\{ \frac{:n \wedge \neg p}{n} \,\middle|\, c \in C, n = N(c), \neg p \in c, M \not\models p \right\} \\
& \cup \left\{ \frac{:n \wedge \neg p'}{n} \,\middle|\, c \in C, n = N(c), p \in c, M \models p \right\}.
\end{aligned}
$$

Define $E = Cn(\{\phi|:\psi/\phi \in A\})$. To verify that $E$ is an extension of $\langle D, \emptyset \rangle$ it suffices to check that for every $:\phi/\psi \in A$, $\neg\phi \notin E$, and for every $:\phi/\psi \in D \backslash A$, $\neg\phi \in E$, and this is straightforward. Hence $\langle D, \emptyset \rangle \not\models_c false$.

Assume that $\langle D, \emptyset \rangle \not\models_c false$; that is, there is an extension $E$ of $\langle D, \emptyset \rangle$ such that $false \notin E$. Let $M$ be a model such that for all propositional variables $p$, $M \models p$ if and only if $p \in E$. We show that $M \models C$. Because $false \notin E$, $:false \wedge \neg n/false \notin GD(E, \Delta)$ for all $n$ such that $n = N(c)$ for some $c \in C$. Hence $n \in E$ for all such $n$. Hence for every $n$, there is $:n \wedge \neg p/n \in GD(E, \Delta)$ or $:n \wedge \neg p'/n \in GD(E, \Delta)$. Hence for every clause $l_1 \vee \cdots \vee l_n \in C$, $p \notin E$ for some $\neg p \in \{l_1, \ldots, l_n\}$, or $p' \notin E$ for some $p \in \{l_1, \ldots, l_n\}$. In the first case by definition $M \models \neg p$. In the second case $p \in E$ and hence by definition $M \models p$. Hence every clause in $C$ is true in $M$, and $C$ is satisfiable. □

Ben-Eliyahu and Dechter (1996) show that testing $\models_c$ for a class of default theories that subsumes all classes in the Kautz and Selman and Stillman hierarchy that have a 2-literal objective part is co-NP-complete. Hence the problems in Theorems 4.12, 4.13 and 4.14 are in co-NP, and consequently co-NP-complete. The Ben-Eliyahu and Dechter result, however, has no direct implications on the complexity of the prioritized versions of these problems.

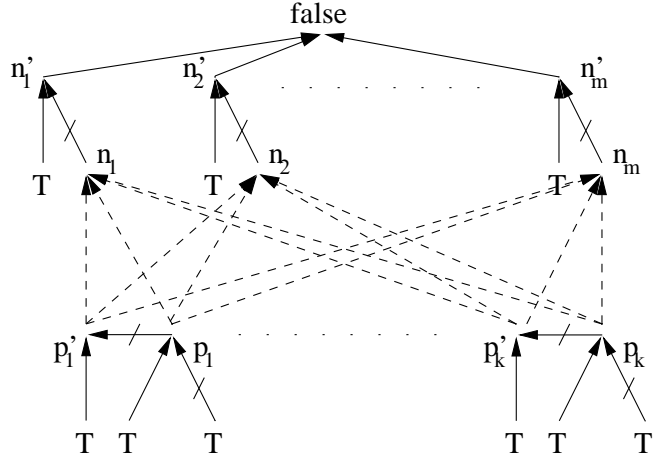

Figure 2: A translation from satisfiability to prioritized default logic

**Theorem 4.15** *Both testing $\Delta \models_{\mathcal{P}}^{B} l$ and $\Delta \models_{\mathcal{P}}^{BH} l$ for normal unary default theories $\Delta$, strict partial orders $\mathcal{P}$ and literals $l$, is co-NP-hard.*

*Proof:* We give the proof for preferred$^{B}$ extensions only. The proof for preferred$^{BH}$ extensions is similar. The proof is by reduction from propositional satisfiability to the complement of the problem. Let $C = \{c_1, \dots, c_m\}$ be a set of propositional clauses and $P$ the set of propositional variables occurring in $C$. Let $N$ be an injective function that maps each clause $c \in C$ to a propositional variable $n = N(c)$ such that $n \notin P$. Define the default theory $\Delta = \langle D, \emptyset \rangle$ and priorities $\mathcal{P}$ on $D$ as follows.

$$D_1 = \left\{ \frac{:p}{p} \,\Big|\, p \in P \right\} \cup \left\{ \frac{:\neg p}{\neg p} \,\Big|\, p \in P \right\}$$

$$D_2 = \left\{ \frac{:p'}{p'} \,\Big|\, p \in P \right\} \cup \left\{ \frac{p:\neg p'}{\neg p'} \,\Big|\, p \in P \right\}$$

$$D_3 = \left\{ \frac{p:n}{n} \,\Big|\, p \in c \in C, n = N(c) \right\} \cup \left\{ \frac{p':n}{n} \,\Big|\, \neg p \in c \in C, n = N(c) \right\}$$

$$\cup \left\{ \frac{:n'}{n'} \,\Big|\, c \in C, n = N(c) \right\} \cup \left\{ \frac{n:\neg n'}{\neg n'} \,\Big|\, c \in C, n = N(c) \right\}$$

$$\cup \left\{ \frac{n':false}{false} \,\Big|\, c \in C, n = N(c) \right\}$$

$$D = D_1 \cup D_2 \cup D_3$$

$$\mathcal{P} = \left\{ \left\langle \frac{p:\neg p'}{\neg p'}, \frac{:p'}{p'} \right\rangle \,\Big|\, p \in P \right\} \cup (D_1 \times (D_2 \cup D_3))$$

Priorities are needed to guarantee that $p \in E$ if and only if $p' \notin E$, and total priorities cannot be used because we cannot restrict to those models that correspond to preferred extensions (with respect to some ordering on the variables), as they are not necessarily models of $C$ even if $C$ is satisfiable. The default theory is depicted in Figure 2. Defaults $p{:}q/q$ are shown as arrows $p \to q$, and defaults $p{:}\neg q/\neg q$ as broken arrows $p \nrightarrow q$. Only some of defaults $p{:}n/n$ and $p'{:}n/n$ for propositions $p \in P$ and $n, n = N(c)$ for $c \in C$, are in $D$,

439

and therefore they are shown as dashed arrows. We claim that $C$ is satisfiable if and only if there is a $\mathcal{P}$-preferred$^B$ extension of $\Delta$ that does not contain *false*.

Assume that $C$ is satisfiable; that is, there is a model $M$ such that $M \models C$. We show that there is a $\mathcal{P}$-preferred$^B$ extension $E$ of $\Delta$ such that *false* $\notin E$. Let $E = Cn(\{p \in P | M \models p\} \cup \{p' | p \in P, M \not\models p\} \cup \{\neg p | p \in P, M \not\models p\} \cup \{\neg p' | p \in P, M \models p\} \cup \{n | c \in C, n = N(c)\} \cup \{\neg n' | c \in C, n = N(c)\})$. Let $\mathcal{T}$ be a strict total order on $D$ such that $\mathcal{P} \subseteq \mathcal{T}$ and for all $p \in P$, $:p/p\mathcal{T}:\neg p/\neg p$ if $M \models p$ and $:\neg p/\neg p\mathcal{T}:p/p$ otherwise, and $n:\neg n'/\neg n'\mathcal{T}:n/n$ for all $c \in C$, $n = N(c)$. It is straightforward to verify that $E$ is a $\mathcal{P}$-preferred$^B$ extension of $\Delta$ generated by $\mathcal{T}$. Clearly *false* $\notin E$.

Assume that $E$ is a $\mathcal{P}$-preferred$^B$ extension such that *false* $\notin E$. Let $M$ be a model such that for all $p \in P$, $M \models p$ if and only if $p \in E$. Because *false* $\notin E$, no default $n':false/false$ is applied in $E$, where $n = N(c)$ for some $c \in C$. Therefore $n' \notin E$ for all $n = N(c)$ such that $c \in C$. Therefore $\neg n' \in E$ and $n \in E$ for all such $n$. Hence a disjunct $p$ is in $E$ or disjunct $p'$ is in $E$ for a disjunct $\neg p$ for every $c \in C$. In the first case by definition of $M$, $M \models c$. In the second case $p \notin E$, because otherwise $\neg p'$ would be in $E$ as $p:\neg p'/\neg p'\mathcal{P}:p'/p'$. Hence $M \models \neg p$ and $M \models c$. Because this holds for all $c \in C$, finally $M \models C$. $\qquad\square$

## 5. Lexicographic Prioritized Default Logic

A definition of prioritized default logic is used by Rintanen (1999). This definition is based on an earlier one for autoepistemic logic (Rintanen, 1994). The priority mechanism uses lexicographic comparison and the preferred extensions in this approach do not in general coincide with the preferred extensions in the prioritized default logics discussed in Section 4. Lexicographic comparison has earlier been used in the context of nonmonotonic reasoning by several researchers (Lifschitz, 1985; Geffner & Pearl, 1992; Ryan, 1992). Comparing two extensions is based on whether the defaults are generating defaults of the extensions, that is, whether their prerequisites belong to the extension and the negations of the justifications do not belong to the extension. We say that a generating default of an extension is *applied* in the extension.

**Definition 5.1 (Application)** *A default* $\alpha:\beta_1, \ldots, \beta_n/\gamma$ *is applied in* $E \subseteq \mathcal{L}$ *if* $E \models \alpha$ *and* $\{\neg\beta_1, \ldots, \neg\beta_n\} \cap Cn(E) = \emptyset$. *This is denoted by* $appl(\alpha:\beta_1, \ldots, \beta_n/\gamma, E)$.

We abbreviate $appl(\delta, E)$ and not $appl(\delta, E')$ by the notation $appl(\delta, E, E')$.

**Definition 5.2 (Preferredness)** *Let* $\Delta = \langle D, W \rangle$ *be a default theory and* $\mathcal{P}$ *a strict partial order on* $D$. *Let* $E$ *be an extension of* $\Delta$. *Then* $E$ *is a* $\mathcal{P}$-preferred$^L$ *extension of* $\Delta$ *if there is a strict total order* $\mathcal{T}$ *on* $D$ *such that* $\mathcal{P} \subseteq \mathcal{T}$ *and for all extensions* $E'$ *of* $\Delta$ *and* $\delta \in D$,

$appl(\delta, E', E)$ *implies that for some* $\epsilon \in D, \epsilon\mathcal{T}\delta$ *and* $appl(\epsilon, E, E')$.

*Such a strict total order is a* $\Delta, \mathcal{P}$-ordering *for* $E$.

Our investigation on lexicographic prioritization in default reasoning was motivated by earlier work on the topic (Tan & Treur, 1992; Baader & Hollunder, 1995). These definitions of priorities for default logic are procedural, as they are given as extensions of (nondeterministic) decision procedures for default logic. This procedural nature of prioritization is sensitive to the lengths of sequences of defaults involved in deriving certain facts: among two conflicting defaults the one with the lower priority may become applied solely because the sequence of defaults needed for deriving its prerequisite is shorter (Brewka & Eiter, 1998). Approaches to prioritizing defaults that are based on lexicographic comparison (Lifschitz, 1985; Brewka, 1989; Geffner & Pearl, 1992; Ryan, 1992) do not exhibit that kind of behavior.

Lexicographic comparison has properties that are favorable from the point of view of knowledge representation. Extensions of a default theory are possible interpretations of the default theory, representing different ways of resolving the conflicts between default rules. Priorities express the plausibility of different ways of resolving the conflicts, and consequently act as an implicit representation of preferences between the extensions. One useful property of lexicographic comparison is that every finite default theory has at least one preferred extension whenever it has at least one Reiter extension. It does not seem plausible that the priorities could contain information that indicates that none of the extensions is a plausible meaning of the default theory. Another useful property is that every extension of a default theory is a $\mathcal{P}$-preferred$^L$ extension for a suitably chosen $\mathcal{P}$. In other words, the way priorities are used in ranking the extensions should not per se rule out the possibility that a certain extension is preferred or that it is the unique preferred extension.

A distinguishing difference between the lexicographic prioritized default logic and other work on priorities in default logic is that the highest priority default – if there is one – is applied in all preferred$^L$ extensions if there is an extension where it is applied. Preferred$^B$ and preferred$^{BH}$ extensions do not always apply the applicable highest priority defaults.

**Example 5.1 (Brewka, 1994)** Consider the default theory $\Delta = \langle D, W \rangle$ where $W = \{a\}$ and $D = \{b:c/c, a:\neg c/\neg c, a:b/b\}$. Define the relation

$$\mathcal{P} = \left\{ \left\langle \frac{b:c}{c}, \frac{a:\neg c}{\neg c} \right\rangle, \left\langle \frac{a:\neg c}{\neg c}, \frac{a:b}{b} \right\rangle, \left\langle \frac{b:c}{c}, \frac{a:b}{b} \right\rangle \right\}.$$

According to Definitions 4.2 and 4.3 the sets $E_i$ are as follows, and the unique $\mathcal{P}$-preferred$^B$ and $\mathcal{P}$-preferred$^{BH}$ extension of $\Delta$ is $E = \bigcup_{i \geq 1} Cn(E_i)$.

$$
\begin{aligned}
E_0 &= \{a\} \\
E_1 &= \{a, \neg c\} \\
E_2 &= \{a, \neg c, b\} \\
E_i &= E_2 \text{ for all } i \geq 3
\end{aligned}
$$

In other words, initially the highest priority default $b:c/c$ is not applicable, and hence the second default $a:\neg c/\neg c$ is applied, and $\neg c$ is obtained. The highest priority default is still not applicable, and hence the third default $a:b/b$ is applied, and $b$ is obtained. Now $b:c/c$ were applicable if the contradicting default $a:\neg c/\neg c$ would not have been applied first. $\square$

The application of the highest priority default – whenever possible – would seem a useful declarative property for nonmonotonic reasoning with priorities. The satisfaction of this property leads to lexicographic prioritization. Also, the behavior of lexicographic prioritized default logic is more consistent for default theories with normal defaults $\alpha:\beta/\beta$ and closely related default theories with prerequisite-free normal defaults :$\alpha \to \beta / \alpha \to \beta$. The latter defaults allow reasoning by contraposition with the implications, and the former do not, but otherwise they represent related patterns of reasoning. The logics by Brewka and by Baader and Hollunder order extensions lexicographically for the latter kind of default theories, but not for the former.

**Definition 5.3** *The consequence relation $\models^L$ is defined by $\Delta \models^L_{\mathcal{P}} \phi$ if and only if the formula $\phi$ is in all $\mathcal{P}$-preferred$^L$ extensions of $\Delta$.*

**Example 5.2** Let $\langle D, W \rangle$ be a default theory where $D = \{p{:}q/q, p{:}\neg r/\neg r, q{:}r/r\}$ and $W = \{p\}$. Let $\mathcal{P} = \{\langle p{:}\neg r/\neg r, q{:}r/r \rangle\}$ be a strict partial order on $D$. The default theory has two extensions, $E_1 = Cn(\{p, q, \neg r\})$ where the defaults $p{:}\neg r/\neg r$ and $p{:}q/q$ are applied, and $E_2 = Cn(\{p, q, r\})$ where $p{:}q/q$ and $q{:}r/r$ are applied. These extensions and all strict total orders $\mathcal{T}$ on $D$ such that $\mathcal{P} \subseteq \mathcal{T}$ are depicted below. The most significant defaults are the lowest. The symbol $\bullet$ signifies that the default is applied and $\circ$ that it is not applied.

$$
\begin{array}{ccc}
& \begin{array}{cc}E_1 & E_2\end{array} & \\
\dfrac{p:q}{q} & \begin{array}{cc}\bullet & \bullet\end{array} & \\
\dfrac{q:r}{r} & \begin{array}{cc}| & \circ\end{array} & \\
\dfrac{p:\neg r}{\neg r} & \begin{array}{cc}\bullet & |\end{array} &
\end{array}
\qquad
\begin{array}{ccc}
& \begin{array}{cc}E_1 & E_2\end{array} & \\
\dfrac{q:r}{r} & \begin{array}{cc}\circ & \bullet\end{array} & \\
\dfrac{p:q}{q} & \begin{array}{cc}| & \bullet\end{array} & \\
\dfrac{p:\neg r}{\neg r} & \begin{array}{cc}\bullet & \circ\end{array} &
\end{array}
\qquad
\begin{array}{ccc}
& \begin{array}{cc}E_1 & E_2\end{array} & \\
\dfrac{q:r}{r} & \begin{array}{cc}\circ & \bullet\end{array} & \\
\dfrac{p:\neg r}{\neg r} & \begin{array}{cc}| & |\end{array} & \\
\dfrac{p:q}{q} & \begin{array}{cc}| & |\end{array} &
\end{array}
$$

The extension $E_1$ is a $\mathcal{P}$-preferred$^L$ extension because the leftmost strict total order $\mathcal{T}_1$ is a $\Delta, \mathcal{P}$-ordering for $E_1$: $q{:}r/r$ is the only default $\delta$ such that appl$(\delta, E_2, E_1)$, and appl$(p{:}\neg r/\neg r, E_1, E_2)$ and $p{:}\neg r/\neg r \mathcal{T}_1 q{:}r/r$. The extension $E_2$ is not a $\mathcal{P}$-preferred$^L$ extension because none of the three strict total orders $\mathcal{T}_1, \mathcal{T}_2, \mathcal{T}_3$ is a $\Delta, \mathcal{P}$-ordering for $E_2$: for all $i \in \{1, 2, 3\}$, there is the default $p{:}\neg r/\neg r$ such that appl$(p{:}\neg r/\neg r, E_1, E_2)$ and there is no default $\delta$ such that $\delta \mathcal{T}_i p{:}\neg r/\neg r$ and appl$(\delta, E_2, E_1)$. $\square$

**Lemma 5.4** *Let $\Delta = \langle D, W \rangle$ be a default theory where $D$ is finite, and let $\mathcal{P}$ be a strict total order on $D$. Let $\Delta$ have at least one extension. Then there is exactly one $\mathcal{P}$-preferred$^L$ extension of $\Delta$.*

*Proof:* We show that there is an extension $E$ of $\Delta$ such that $\mathcal{P}$ is the $\Delta, \mathcal{P}$-ordering for $E$. Let $\delta_1, \ldots, \delta_n$ be the ordering $\mathcal{P}$ of $D$. Define $D_i = \{\delta_1, \ldots, \delta_i\}$ for all $i \in \{0, \ldots, n\}$. Define for all $i \in \{1, \ldots, n-1\}$,

$$
\begin{aligned}
X_0 &= \{E \subseteq \mathcal{L} | E \text{ is an extension of } \Delta\}, \text{ and} \\
X_{i+1} &= \begin{cases} \{E \in X_i | \text{appl}(\delta_{i+1}, E)\} & \text{if appl}(\delta_{i+1}, E) \text{ for some } E \in X_i, \\ X_i & \text{otherwise.} \end{cases}
\end{aligned}
$$

*Induction hypothesis*: For $j \in \{0, \ldots, i\}$, (1) the set $X_j$ is non-empty, (2) for all $E \in X_j$ and $E' \in X_j$ and $\delta \in D_j$, appl$(\delta, E)$ iff appl$(\delta, E')$, and (3) for all $E \in X_j$ and $E' \in X_0 \backslash X_j$ there is $\delta \in D_j$ such that appl$(\delta, E, E')$ and there is no $\delta' \in D$ such that $\delta' \mathcal{P} \delta$ and appl$(\delta', E', E)$.

The proofs of both the base case and the inductive case are straightforward.

The claim of the lemma is obtained from the facts established in the induction proof as follows. By (1) the set $X_n$ is non-empty. By (2) and Theorems 2.4 and 2.5 in (Reiter, 1980)

```
PROCEDURE extension(D, W, E);
  E' := W;
  REPEAT
    E'' := E';
    FOR EACH α:β₁,...,βₙ/γ ∈ D DO
      IF ⟨E', α⟩ ∈ CN and ⟨E, ¬β⟩ ∉ CN for all β ∈ {β₁,...,βₙ}
        THEN E' := E' ∪ {γ}
    END
  UNTIL E' = E'';
  IF ⟨E, φ⟩ ∈ CN for all φ ∈ E' and ⟨E', φ⟩ ∈ CN for all φ ∈ E
    THEN RETURN true ELSE RETURN false
END
```

Figure 3: A procedure for recognizing extensions

$|X_n| \leq 1$. Hence $X_n$ is a singleton $\{E\}$. Let $E'$ be any extension of $\Delta$. Assume that there is $\delta' \in D$ such that $\mathrm{appl}(\delta', E', E)$. Hence $E' \neq E$ and $E' \in X_0 \setminus X_n$. Now by (3) there is $\delta \in D_n = D$ such that $\mathrm{appl}(\delta, E, E')$ and there is no $\delta''$ such that $\delta'' \mathcal{P} \delta$ and $\mathrm{appl}(\delta'', E', E)$. Therefore not $\delta' \mathcal{P} \delta$, and because $\mathcal{P}$ is a strict total order, it is the case that $\delta \mathcal{P} \delta'$. Because this holds for all $\delta' \in D$ and all extensions $E'$ of $\Delta$, $\mathcal{P}$ is a $\Delta$, $\mathcal{P}$-ordering for $E$. Therefore $E$ is a $\mathcal{P}$-preferred$^L$ extension of $\Delta$. Let $E'$ be any extension such that $E \neq E'$. Now $E \in X_n$ and $E' \notin X_n$, and therefore by (3) there is $\delta \in D$ such that $\mathrm{appl}(\delta, E, E')$ and there is no $\delta' \in D$ such that $\delta' \mathcal{P} \delta$ and $\mathrm{appl}(\delta', E', E)$. Hence $E$ is the only $\mathcal{P}$-preferred$^L$ extension of $\Delta$. □

## 5.1 Complexity in the General Case

The language that corresponds to the consequence relation $\models$ of the classical propositional logic is denoted by CN and is defined as the set of pairs $\langle \Sigma, \phi \rangle \in 2^{\mathcal{L}} \times \mathcal{L}$ such that $\Sigma \models \phi$. Some of the complexity results use the procedure in Figure 3 that is directly based on the semiconstructive definition of extensions given in Theorem 3.2.

**Lemma 5.5** *For the procedure in Figure 3, the call extension$(D, W, E)$ returns* true *if and only if $Cn(E)$ is an extension of the default theory $\langle D, W \rangle$. Excluding the tests of membership in CN the procedure runs in polynomial time on the size of $\langle D, W \rangle$.*

Our decision procedure for the lexicographic prioritized default logic is based on a reduction to the language ENC, which in turn is reducible to CN in nondeterministic polynomial time. There are three kinds of questions ENC can answer: is the logical closure of a set of formulae an extension of a default theory, is a strict total order a $\Delta$, $\mathcal{P}$-ordering for an extension, and is a formula a logical consequence of a set of formulae. The language ENC $\subseteq \{0, 1, 2\} \times 2^{\mathcal{D}} \times 2^{\mathcal{L}} \times 2^{\mathcal{L}} \times (2^{\mathcal{D} \times \mathcal{D}} \cup \mathcal{L})$ is defined as the set of quintuples

- $\langle 0, D, W, E, \mathcal{T} \rangle$ where $D$ is a set of defaults, $W \subseteq \mathcal{L}$ is a set of formulae, and $Cn(E)$ is an extension of $\langle D, W \rangle$,

*PROCEDURE* ENC($n, D, W, E, \mathcal{T}$);
  *CASE n OF*
  0: *IF* extension($D, W, E$) *THEN* accept *ELSE* reject
  1: guess nondeterministically a subset $E'$ of conclusions of $D$;
    $E' := E' \cup W$;
    *IF* not extension($D, W, E'$) or not extension($D, W, E$)
      *THEN* reject
    *ELSE IF* compare($E, E', \mathcal{T}$) = true *THEN* reject
        *ELSE* accept
  2: *IF* $\langle E, \mathcal{T} \rangle \in$ CN *THEN* accept *ELSE* reject
  otherwise: reject
*END*

*PROCEDURE* applied($\alpha{:}\beta_1, \ldots, \beta_n / \gamma, E$);
  *IF* $\langle E, \alpha \rangle \in$ CN and $\langle E, \neg\beta \rangle \notin$ CN for all $\beta \in \{\beta_1, \ldots, \beta_n\}$
  *THEN RETURN* true *ELSE RETURN* false
*END*

*PROCEDURE* compare($E, E', \mathcal{T}$);
  let $\mathcal{T}$ be the ordering $\delta_0 \mathcal{T} \delta_1 \mathcal{T} \cdots \mathcal{T} \delta_n$;
  *FOR* i := 0 *TO* n *DO*
    *IF* not applied($\delta_i, E$) and applied($\delta_i, E'$) *THEN RETURN* false;
    *IF* applied($\delta_i, E$) and not applied($\delta_i, E'$) *THEN RETURN* true
  *END*
  *RETURN* true
*END*

Figure 4: A decision procedure for ENC

- $\langle 1, D, W, E, \mathcal{T} \rangle$ where $D$ is a set of defaults, $W \subseteq \mathcal{L}$ is a set of formulae, $E$ is a set of formulae, $\mathcal{T}$ is a strict total order on $D$, and $\mathcal{T}$ is not a $\langle D, W \rangle, \mathcal{T}$-ordering for $Cn(E)$, and

- $\langle 2, D, W, E, \phi \rangle$ where $\langle E, \phi \rangle \in$ CN.

**Lemma 5.6** *ENC is Turing reducible to CN in nondeterministic polynomial time.*

*Proof:* The language ENC is accepted by a nondeterministic Turing machine given as the procedure in Figure 4. The executions of the procedure have a polynomial length. $\quad\Box$

**Theorem 5.7** *The complement of the problem of testing whether a formula belongs to all preferred$^L$ extensions of a default theory is Turing reducible to ENC in nondeterministic polynomial time.*

*PROCEDURE* co-cautious($D, W, \mathcal{P}, \phi$);
    guess nondeterministically a subset $E$ of conclusions of $D$;
    guess nondeterministically a strict total order $\mathcal{T}$ on $D$;
    *IF* $\mathcal{P} \not\subseteq \mathcal{T}$ *THEN* reject;
    *IF* $\langle 0, D, W, E \cup W, \emptyset \rangle \in$ ENC
       *AND* $\langle 1, D, W, E \cup W, \mathcal{T} \rangle \notin$ ENC
       *AND* $\langle 2, \emptyset, \emptyset, E \cup W, \phi \rangle \notin$ ENC
    *THEN* accept
    *ELSE* reject
*END*

Figure 5: A decision procedure for prioritized cautious reasoning

*Proof:* The question is answered by a Turing machine described by the procedure in Figure 5. The Turing machine uses an oracle for the language ENC. First the machine guesses a candidate extension $Cn(E \cup W)$ of $\langle D, W \rangle$ and a strict total order $\mathcal{T}$ on $D$ that extends the relation $\mathcal{P}$. Then the machine verifies that $Cn(E \cup W)$ is in fact an extension (the first consultation of the oracle for ENC), that $\mathcal{T}$ is a $\langle D, W \rangle$, $\mathcal{P}$-ordering for $Cn(E \cup W)$ (the second consultation), and that $\phi \notin Cn(E \cup W)$ (the third consultation.) If all these verifications succeed, the Turing machine accepts. Because the guesses of $E$ and $\mathcal{T}$ are nondeterministic, the machine accepts whenever there is an extension that contains $\phi$ and has a $\langle D, W \rangle$, $\mathcal{P}$-ordering.

The only parts of the procedure that do not run in constant time are the nondeterministic guessing of the set $E$, guessing of a strict total order $\mathcal{T}$, and verifying that it extends $\mathcal{P}$. Because the sizes of $E$ and $\mathcal{T}$ are polynomial in $\langle D, W \rangle$, the number of steps needed to guess them is polynomial in $\langle D, W \rangle$. Verification of $\mathcal{P} \subseteq \mathcal{T}$ takes polynomial time. $\qquad\square$

We give an upper and a lower bound for the location of the decision problem in the polynomial hierarchy. Because ENC is Turing reducible to CN in nondeterministic polynomial time, the following lemma is immediate.

**Lemma 5.8** *The language ENC is in* $\Sigma_2^p$.

By Theorem 5.7 we get an upper bound for the location of the decision problem in the polynomial hierarchy. The result is immediate by the definition of the polynomial hierarchy.

**Theorem 5.9** *The problem of whether a formula belongs to all preferred$^L$ extensions of a default theory is in* $\Pi_3^p$.

Next we show that the decision problem is $\Delta_3^p$-hard, thereby obtaining a lower bound on the location in the polynomial hierarchy. This result is based on a theorem by Krentel (1992) that identifies $\Delta_k^p$-complete problems for $k \geq 1$. The theorem concerns lexicographically maximal valuations of quantified Boolean formulae, and for the prefix $\exists\forall$ it is as follows.

**Theorem 5.10 (Krentel, 1992)** *The problem of computing a given component of* $X^1$ *is* $\Delta_3^p$-*complete, where* $X^1$ *is defined as follows.*

Maximum 2-quantified formula
Instance: *Boolean formula $C[X^1, X^2]$ where $X^i$ is an abbreviation for an $n$-tuple of Boolean variables that corresponds to a valuation of $x_1^i, \ldots, x_n^i$.*
Output: *Lexicographically maximal $X^1 \in \{0, 1\}^n$ that satisfies*

$$\exists X^1 \forall X^2 (C[X^1, X^2] = 1).$$

The following lemma points out a connection between quantified Boolean formulae with a universal quantifier as the outermost quantifier and the logical consequence relation in propositional logic.

**Lemma 5.11** *Let $X$ be a set containing exactly one of $x$ and $\neg x$ for every $x \in \{x_1^1, \ldots, x_n^1\}$. Let $X^1 = \langle x_1'^1, \ldots, x_n'^1 \rangle$ where $x_i'^1 = 1$ if $x_i^1 \in X$ and $x_i'^1 = 0$ otherwise. Then $X^1$ satisfies $\forall X^2 (C[X^1, X^2])$ if and only if $X \models C[X^1, X^2]$.*

*Proof:* Assume that $X^1$ satisfies $\forall X^2 (C[X^1, X^2])$. Hence for the choice of truth values for $x_1^1, \ldots, x_n^1$ indicated by $X^1$ and any choice of truth values for $x_1^2, \ldots, x_n^2$, $C[X^1, X^2]$ is true. Now let $M$ be any model such that $M \models X$. Hence $M$ assigns the same truth values to $x_1^1, \ldots, x_n^1$ as $X^1$. Now $M \models C[X^1, X^2]$, and by the definition of logical consequence $X \models C[X^1, X^2]$.

Assume that $X \models C[X^1, X^2]$. Hence for all models $M$ such that $M \models X$ and $M$ assigns any truth-values to $x_1^2, \ldots, x_n^2$, $M \models C[X^1, X^2]$. By definition of quantified Boolean formulae, for $X^1$ and any $X^2$ the formula $C[X^1, X^2]$ is true, that is, $X^1$ satisfies $\forall X^2 (C[X^1, X^2])$. $\square$

The proof of the next lemma uses the same translation of quantified Boolean formulae to default theories as Gottlob's proof of $\Pi_2^p$-hardness of cautious reasoning in Reiter's default logic (Gottlob, 1992).

**Lemma 5.12** *Computing a given component of $X^1$ in Maximum 2-quantified formula is polynomial time many-one reducible to the problem of testing whether a formula belongs to all $\mathcal{T}$-preferred$^L$ extensions of a default theory for strict total orders $\mathcal{T}$.*

*Proof:* We construct a default theory $\Delta = \langle D, \emptyset \rangle$ and define a strict partial order $\mathcal{P}$ on $D$ so that the $k$th component of $X^1$ is 1 if and only if $\Delta \models_{\mathcal{P}}^L x_k^1$. The value of a given component of $X^1$ in Maximum 2-quantified formula is defined only if $\exists X^1 \forall X^2 (C[X^1, X^2])$ is satisfiable, so assume it is. Let $a$ be a propositional variable that is not in $\{x_1^1, \ldots, x_n^1, x_1^2, \ldots, x_n^2\}$. Define

$$D = \left\{ \frac{: x_1^1}{x_1^1}, \frac{: \neg x_1^1}{\neg x_1^1}, \ldots, \frac{: x_n^1}{x_n^1}, \frac{: \neg x_n^1}{\neg x_n^1}, \frac{C[X^1, X^2] : a}{a} \right\}, \text{ and}$$

$$\begin{aligned} \mathcal{P} = &\left\{ \frac{C[X^1, X^2] : a}{a} \right\} \times \left( D \backslash \left\{ \frac{C[X^1, X^2] : a}{a} \right\} \right) \\ &\cup \left\{ \left\langle \frac{: x_i^1}{x_i^1}, \frac{: x_j^1}{x_j^1} \right\rangle \middle| 1 \leq i < j \leq n \right\} \\ &\cup \left\{ \frac{: x_1^1}{x_1^1}, \ldots, \frac{: x_n^1}{x_n^1} \right\} \times \left\{ \frac{: \neg x_1^1}{\neg x_1^1}, \ldots, \frac{: \neg x_n^1}{\neg x_n^1} \right\}. \end{aligned}$$

Let $\mathcal{T}$ be any strict total order on $D$ that extends $\mathcal{P}$. The sizes of the default theory $\langle D, \emptyset \rangle$ and the strict total order $\mathcal{T}$ are linearly proportional to the size of $C[X^1, X^2]$, and they can be constructed in polynomial time.

First we show that for each extension $E$ of $\langle D, \emptyset \rangle$ that contains $a$ there is $X^1$ such that $X^1$ satisfies $\forall X^2(C[X^1, X^2])$ and for all $i \in \{1, \ldots, n\}$, $p_i(X^1) = 1$ iff $x_i^1 \in E$ (the function $p_i$ selects the $i$th element of an $n$-tuple.) Let $E$ be an extension of $\langle D, \emptyset \rangle$ and $a \in E$. Because defaults in $D$ are normal, $E$ is consistent. Hence $\mathrm{appl}(C[X^1, X^2]{:}a/a, E)$ because the only default where $a$ occurs in the conclusion is $C[X^1, X^2]{:}a/a$. Hence $C[X^1, X^2] \in E$; that is, $C[X^1, X^2]$ is a logical consequence of the conclusions of generating defaults of $E$. Because $a$ does not occur in $C[X^1, X^2]$, $X \models C[X^1, X^2]$ for $X = E \cap \{x_1^1, \neg x_1^1, \ldots, x_n^1, \neg x_n^1\}$. By Lemma 5.11 $X^1$, defined as $p_i(X^1) = 1$ if $x_i^1 \in X$ and $p_i(X^1) = 0$ otherwise, satisfies $\forall X^2(C[X^1, X^2])$.

Then we show that for each $X^1$ that satisfies $\forall X^2(C[X^1, X^2])$, there is an extension $E$ of $\langle D, \emptyset \rangle$ such that $a \in E$ and $p_i(X^1) = 1$ iff $x_i^1 \in E$. Assume that $X^1$ satisfies $\forall X^2(C[X^1, X^2])$. Let $X = \{x_i^1 | p_i(X^1) = 1, 1 \le i \le n\} \cup \{\neg x_i^1 | p_i(X^1) = 0, 1 \le i \le n\}$. It is straightforward to verify that $E$ is an extension of $\langle D, \emptyset \rangle$. Because the extension $E$ is consistent, $\mathrm{appl}(C[X^1, X^2]{:}a/a, E)$. Because $C[X^1, X^2]{:}a/a$ is the $\mathcal{P}$-least default in $D$, all $\mathcal{T}$-preferred$^L$ extensions apply $C[X^1, X^2]{:}a/a$.

The counterpart of the $\mathcal{T}$-preferred$^L$ extension is the lexicographically maximal $X^1$, and vice versa. This is by the construction of $\mathcal{T}$, where the defaults ${:}x_1^1/x_1^1, \ldots, {:}x_n^1/x_n^1$ are ordered as ${:}x_i^1/x_i^1 \mathcal{T} {:} x_j^1/x_j^1$ whenever $i < j$. Therefore $p_i(X^1) = 1$ for the maximal $X^1$ satisfying $\forall X^2(C[X^1, X^2])$ iff $x_i^1$ belongs to the $\mathcal{T}$-preferred$^L$ extension of $\langle D, \emptyset \rangle$. □

Notice that Lemma 5.12 cannot be reconstructed in the prioritized default logics by Brewka and by Baader and Hollunder because the construction is based on a highest priority default that becomes applicable only after a number of lower priority defaults have been applied, and in these cases those logics do not work like the lexicographic prioritized default logic. While testing $\models_{\mathcal{T}}^L$ for strict total orders $\mathcal{T}$ is higher in the polynomial hierarchy than cautious reasoning in unprioritized default logic, testing $\models_{\mathcal{T}}^B$ and $\models_{\mathcal{T}}^{BH}$ is one level lower.

**Theorem 5.13** *For default theories $\Delta = \langle D, W \rangle$, strict total orders $\mathcal{T}$ on $D$, and formulae $\phi$, the problem of testing $\Delta \models_{\mathcal{T}}^L \phi$ is $\Delta_3^p$-hard.*

*Proof:* All problems in $\Delta_3^p$ are polynomial time many-one reducible to MAXIMUM 2-QUANTIFIED FORMULA, which by Lemma 5.12 is polynomial time many-one reducible to our problem. □

Testing $\models_{\mathcal{T}}^L$ for strict total orders $\mathcal{T}$ is in $\Delta_3^p$. The proof of this fact is based on computing the unique preferred extensions step by step, at each step determining whether there is an extension in which one of the defaults is applied in addition to those defaults which we have already committed to.

**Lemma 5.14** *For the procedure in Figure 6, the call exists$(D, W, A)$ returns true if and only if there is an extension $E$ of $\langle D, W \rangle$ such that $\mathrm{appl}(\delta, E)$ for all $\delta \in A$. The procedure runs in nondeterministic polynomial time given an NP oracle for CN.*

*Proof:* First the procedure guesses a set of conclusions $G$ that possibly are the conclusions of the generating defaults of the possible extension $Cn(G \cup W)$. The call to extension$(D, W, E)$

*PROCEDURE* exists($D$,$W$,$A$);
   guess a subset $G$ of conclusions of defaults in $D$;
   $E := G \cup W$;
   *IF* not extension($D, W, E$) *THEN RETURN* false;
   *FOR EACH* $\alpha{:}\beta_1, \ldots, \beta_n / \gamma \in A$ *DO*
     *IF* $\langle E, \alpha \rangle \notin$ CN or $\langle E, \neg\beta \rangle \in$ CN for some $\beta \in \{\beta_1, \ldots, \beta_n\}$
     *THEN RETURN* false
   *END*;
   *RETURN* true
*END*

Figure 6: A procedure that tests for the existence of extensions

*PROCEDURE* decide($D$,$W$,$\mathcal{T}$,$\phi$);
   *IF* not exists($D, W, \emptyset$) *THEN RETURN* true;
   $A := \emptyset$;
   let $\delta_1, \ldots, \delta_n$ be the ordering $\mathcal{T}$;
   *FOR* $i := 1$ *TO* $n$ *DO*
     *IF* exists($D, W, A \cup \{\delta_i\}$) *THEN* $A := A \cup \{\delta_i\}$
   *END*
   $E := W \cup \{\gamma | \alpha{:}\sigma / \gamma \in A\}$;
   *IF* $E \models \phi$ *THEN RETURN* true *ELSE RETURN* false
*END*

Figure 7: A decision procedure for total priorities

with $E = G \cup W$ tests if this is indeed the case (Lemma 5.5.) Next the procedure tests whether defaults in $A$ are applied in $Cn(G \cup W)$, and returns true if and only if this is the case. Because the procedure represents a nondeterministic Turing machine, the procedure returns true if and only if it is possible to guess $G$ so that $Cn(G \cup W)$ is an extension where members of $A$ is applied. Hence *exists*($D, W, A$) returns true if and only if there is an extension $E$ of $\langle D, W \rangle$ such that appl($\delta, E$) for all $\delta \in A$.

The procedure runs in non-deterministic polynomial time and uses an NP oracle for propositional satisfiability. Hence the problem solved by it is in $\Sigma_2^p$.      $\square$

**Theorem 5.15** *For default theories* $\Delta = \langle D, W \rangle$*, formulae* $\phi$*, and strict total orders* $\mathcal{T}$ *on* $D$*, the problem of testing* $\Delta \models_{\mathcal{T}}^L \phi$ *is in* $\Delta_3^p$.

*Proof:* We give a decision procedure for the problem that runs in deterministic polynomial time and uses an oracle for a problem that belongs to $\Sigma_2^p$. This demonstrates that the problem is in $\Delta_3^p$. The procedure *decide* is given in Figure 7 and the $\Sigma_2^p$ oracle procedure *exists* is given in Figure 6 and its correctness is stated in Lemma 5.14. The correctness of the procedure *decide* is as follows. Let $\langle D, W \rangle$ be a finite default theory with $|D| = n$, $\mathcal{T}$ a strict total order on $D$, and $\phi$ a formula. Assume that $\langle D, W \rangle$ has no extensions. In this case the first statement in the procedure *decide* returns true which is correct because $\phi$ trivially belongs to all extensions of $\langle D, W \rangle$. Assume that $\langle D, W \rangle$ has at least one extension. The correctness proof in this case proceeds by induction on $i \in \{1, \ldots, n\}$. The value of the program variable $A$ after the $i$th iteration is denoted by $A_i$. Let $\delta_1, \ldots, \delta_n$ be the ordering $\mathcal{T}$. Because $\mathcal{T}$ is a strict total order on $D$, by Lemma 5.4 there is exactly one $\mathcal{T}$-preferred$^L$ extension $E$ of $\langle D, W \rangle$.

*Induction hypothesis*: for all $j \in \{1, \ldots, i\}$, appl$(\delta_j, E)$ if and only if $\delta_j \in A_i$.

*Base case $i = 0$*: Now $A_0 = \emptyset$ and $\{1, \ldots, i\} = \emptyset$. Hence the hypothesis is true.

*Inductive case $i \geq 1$*: We analyze the value $b$ returned by exists$(D, W, A_{i-1} \cup \{\delta_i\})$ by cases. Assume that $b =$ true. Hence $A_i = A_{i-1} \cup \{\delta_i\}$. By the correctness of the procedure *exists*, there is an extension $E'$ such that appl$(\delta, E')$ for all $\delta \in A_{i-1} \cup \{\delta_i\}$. Assume that $E$ is not such an extension. Hence appl$(\delta_i, E', E)$. By induction hypothesis appl$(\delta_j, E)$ iff $\delta_j \in A_{i-1}$ for all $j \in \{1, \ldots, i-1\}$. Now because appl$(\delta, E)$ implies appl$(\delta, E')$ for all $\delta \mathcal{T} \delta_i$ there are no $\delta'' \mathcal{T} \delta_i$ such that appl$(\delta'', E, E')$. Hence $E$ could not be $\mathcal{T}$-preferred$^L$, and the assumption that not appl$(\delta_i, E)$ was wrong. Assume that $b =$ false. As there is no extension $E'$ such that appl$(\delta, E')$ for all $\delta \in A_{i-1} \cup \{\delta_i\}$, and appl$(\delta, E)$ for all $\delta \in A_{i-1}$, not appl$(\delta_i, E)$. Now $A_i = A_{i-1}$, and the induction hypothesis is fulfilled. This finishes the induction proof.

Now $A_n$ is the set of generating defaults of the unique $\mathcal{T}$-preferred$^L$ extension $E$ of $\langle D, W \rangle$. The execution of the program continues at the last statement of the procedure. By Theorem 3.4 $E = Cn(W \cup \{\gamma | \alpha {:} \sigma / \gamma \in A_n\})$. Hence $\phi \in E$ if and only if $W \cup \{\gamma | \alpha {:} \sigma / \gamma \in A_n\} \models \phi$. $\qquad \square$

**Corollary 5.16** *For default theories $\Delta = \langle D, W \rangle$, strict total orders $\mathcal{P}$ on $D$, and formulae $\phi$, the problem of testing $\Delta \models^L_{\mathcal{P}} \phi$ is $\Delta^p_3$-complete.*

*Proof:* Directly by Theorems 5.13 and 5.15. $\qquad \square$

For the problem of membership of formulae all preferred extensions of a default theory without the restriction to total priorities, the membership in $\Delta^p_3$ as well as its $\Pi^p_3$-hardness remain open.

## 5.2 Complexity in Syntactically Restricted Cases

As in Section 4.2, we analyze the complexity of the consequence relation of the lexicographic prioritized default logic under syntactic restrictions.

### 5.2.1 SUMMARY

The results on the boundary between tractability and intractability for $\models^L_{\mathcal{P}}$ with strict total orders $\mathcal{P}$ are summarized in Table 4. In cases where the complexity of the lexicographic prioritized default logic differs from Reiter's default logic, the complexity is set in boldface. Table 5 indicates where the proofs of the complexity results can be found. The impact of priorities on the complexity of the decision problems is strongest in the prerequisite-free normal classes. When the default rules in these classes are totally ordered, the unique preferred extensions can be easily found by applying the default rules in the given order. A more complicated polynomial time decision procedure exists for Horn defaults with 1-literal objective facts (Theorem 5.22). For this class, the existence of an extension that applies a given set of defaults can be determined in polynomial time, and this together with the stepwise construction of preferred extensions given as the procedure in Figure 7 yields a fast decision procedure.

| class of default theories | complexity when clauses in W are | | |
|---|---|---|---|
| | Horn | 2-literal | 1-literal |
| 1  disjunction-free | NP-hard | NP-hard | NP-hard |
| 2  unary | NP-hard | NP-hard | NP-hard |
| 3  disjunction-free ordered | NP-hard | NP-hard | NP-hard |
| 4  ordered unary | NP-hard | NP-hard | NP-hard |
| 5  disjunction-free normal | NP-hard | NP-hard | NP-hard |
| 6  Horn | NP-hard | NP-hard | **PTIME** |
| 7  normal unary | NP-hard | NP-hard | PTIME |
| 8  prerequisite-free | NP-hard | NP-hard | NP-hard |
| 9  prerequisite-free ordered | NP-hard | NP-hard | NP-hard |
| 10  prerequisite-free unary | NP-hard | NP-hard | NP-hard |
| 11  prerequisite-free ordered unary | NP-hard | NP-hard | NP-hard |
| 12  prerequisite-free normal | **PTIME** | **PTIME** | **PTIME** |
| 13  prerequisite-free normal unary | **PTIME** | **PTIME** | PTIME |
| 14  prerequisite-free positive normal unary | **PTIME** | **PTIME** | PTIME |

Table 4: Complexity of the consequence relation $\models^L$ with total priorities

| class of default theories | reference | | |
|---|---|---|---|
| | Horn | 2-literal | 1-literal |
| 1  disjunction-free | 4 $\subseteq$ | 4 $\subseteq$ | 4 $\subseteq$ |
| 2  unary | 4 $\subseteq$ | 4 $\subseteq$ | 4 $\subseteq$ |
| 3  disjunction-free ordered | 5 $\subseteq$ | 5 $\subseteq$ | 5 $\subseteq$ |
| 4  ordered unary | C5.20 | C5.20 | C5.20 |
| 5  disjunction-free normal | C5.18 | C5.18 | C5.18 |
| 6  Horn | 7 $\subseteq$ | 7 $\subseteq$ | T5.22 |
| 7  normal unary | C5.20 | C5.20 | $\subseteq$ 6 |
| 8  prerequisite-free | 11 $\subseteq$ | 11 $\subseteq$ | 11 $\subseteq$ |
| 9  prerequisite-free ordered | 11 $\subseteq$ | 11 $\subseteq$ | 11 $\subseteq$ |
| 10  prerequisite-free unary | 11 $\subseteq$ | 11 $\subseteq$ | 11 $\subseteq$ |
| 11  prerequisite-free ordered unary | C5.20 | C5.20 | C5.20 |
| 12  prerequisite-free normal | T5.21 | T5.21 | T5.21 |
| 13  prerequisite-free normal unary | $\subseteq$ 12 | $\subseteq$ 12 | $\subseteq$ 12 |
| 14  prerequisite-free positive normal unary | $\subseteq$ 12 | $\subseteq$ 12 | $\subseteq$ 12 |

Table 5: References to theorems on the complexity of $\models^L$ with total priorities

### 5.2.2 Intractable Classes with Total Priorities

The next theorems are based on reductions from intractable problems of brave reasoning in Reiter's default logic to prioritized default logic with total priorities.

**Theorem 5.17** *Let $\mathcal{F}$ be a class of formulae and $\mathcal{C}$ a class of finite default theories such that*

- *if $\langle D, W \rangle \in \mathcal{C}$, then $\langle D \cup \{\phi : p/p\}, W \rangle \in \mathcal{C}$ where $\phi \in \mathcal{F}$ and $p$ is a propositional variable that does not occur in $D$ or $W$, and*

- *each member of $\mathcal{C}$ has at least one extension.*

*The problem of testing $\Delta \models_b \phi$ for $\Delta \in \mathcal{C}$ and $\phi \in \mathcal{F}$ is polynomial time many-one reducible to the problem of testing $\Delta' \models_{\mathcal{P}}^L p$ where $\Delta' \in \mathcal{C}$, $p$ is a propositional variable, and $\mathcal{P}$ is a strict total order on defaults in $\Delta'$.*

*Proof:* Let $\langle D, W \rangle \in \mathcal{C}$ be a default theory, $\phi \in \mathcal{F}$ a formula, and $k$ a propositional variable that does not occur in $\langle D, W \rangle$ or $\phi$. Let $\mathcal{P}$ be any strict partial order on $D \cup \{\phi : k/k\}$ such that $\phi : k/k$ is the $\mathcal{P}$-least element. We claim that $\langle D, W \rangle \models_b \phi$ if and only if $\langle D \cup \{\phi : k/k\}, W \rangle \models_{\mathcal{P}}^L k$. This is directly because the highest priority default is applied in all preferred[L] extensions if it is possible to apply it, and $k$ belongs to those extensions if and only if $\phi$ belongs to them. □

As a corollary, together with a theorem that shows the intractability of brave reasoning of the class in Reiter's default logic (Kautz & Selman, 1991), we obtain the following result.

**Corollary 5.18** *Testing $\langle D, W \rangle \models_{\mathcal{P}}^L l$ for strict total orders $\mathcal{P}$, literals $l$, and disjunction-free normal default theories $\langle D, W \rangle$ where $W$ is a set of literals, is NP-hard.*

**Theorem 5.19** *Let $\mathcal{C}$ be a class of finite default theories such that the conclusion of each default is a literal and each member of $\mathcal{C}$ has at least one extension. The problem of testing $\Delta \models_b l$ for $\Delta \in \mathcal{C}$ and literals $l$ is polynomial time many-one reducible to the problem of testing $\Delta \models_{\mathcal{T}}^L l$ where $\mathcal{T}$ is a strict total order on defaults in $\Delta$.*

*Proof:* Let $\Delta = \langle D, W \rangle$ be a default theory in $\mathcal{C}$. We reduce testing $\Delta \models_b l$ to $\Delta \models_{\mathcal{T}}^L l$ as follows. Let $D'$ be the set of defaults with $l$ as the conclusion. Let $\mathcal{T}$ be a strict total order on $D$ such that $D' \times (D \backslash D') \subseteq \mathcal{T}$. We claim that $\Delta \models_b l$ if and only if $\Delta \models_{\mathcal{T}}^L l$.

Assume that $\Delta \not\models_b l$. Because by assumption there is at least one extension of $\Delta$, by Lemma 5.4 there is exactly one $\mathcal{T}$-preferred[L] extension $E$ of $\Delta$. Clearly $l \notin E$. Therefore $\Delta \not\models_{\mathcal{T}}^L l$.

Assume that $\Delta \models_b l$. Then there is an extension $E$ of $\Delta$ such that $l \in E$. If $W \models l$, then clearly $\Delta \models_{\mathcal{T}}^L l$. Assume that $W \not\models l$. Now appl$(\delta, E)$ for some $\delta \in D'$. Assume that $E'$ is an extension of $\Delta$ such that $l \notin E$. Now appl$(\delta, E, E')$. As $\delta' \in D'$ for all $\delta' \mathcal{T} \delta$ and not appl$(\delta', E')$ for all $\delta' \in D'$, there is no $\delta' \mathcal{T} \delta$ such that appl$(\delta', E', E)$. Hence $E'$ is not $\mathcal{T}$-preferred[L]. Therefore $l$ belongs to all $\mathcal{T}$-preferred[L] extensions of $\Delta$ and $\Delta \models_{\mathcal{T}}^L l$. □

The following corollary is obtained with the intractability results of brave reasoning by Kautz and Selman (1991) and Stillman (1990) for the classes mentioned.

**Corollary 5.20** *Testing* $\langle D, W \rangle \models^L_{\mathcal{P}} l$ *for strict total orders $\mathcal{P}$ and literals $l$ in any of the classes of default theories below is NP-hard.*

| $D$ | $W$ |
|---|---|
| *normal unary* | *Horn clauses* |
| *normal unary* | *2-literal clauses* |
| *ordered unary* | *literals* |
| *prerequisite-free ordered unary* | *literals* |

### 5.2.3 Tractable Classes

Next we give a restricted tractable class of prioritized default theories for which the unique $\mathcal{P}$-preferred$^L$ extensions can be computed by using $\mathcal{P}$ as the order in which the application of defaults is attempted. Exhaustive search is avoided because there is no need to retract a decision to apply a default. The theorem is given in a general form for any tractable subset of classical propositional logic, like Horn clauses, 2-literal clauses, or 1-literal clauses. The theorem and the algorithm are part of the folklore of nonmonotonic reasoning.

**Theorem 5.21** *Let $\mathcal{F}$ be a class of formulae such that satisfiability testing for $S \subseteq \mathcal{F}$ takes polynomial time in the size of $S$. Let $\mathcal{C}$ be a class of default theories $\langle D, W \rangle$ such that $D = \{ {:}\phi/\phi | \phi \in F \}$ for some finite set $F \subseteq \mathcal{F}$ and $W \subseteq \mathcal{F}$ is finite. Then for $\Delta \in \mathcal{C}$, $\phi \in \mathcal{F}$, and strict total orders $\mathcal{P}$ on the defaults in $\Delta$, the problem of testing $\Delta \models^L_{\mathcal{P}} \neg\phi$ is solvable in polynomial time on the size of $D \cup W \cup \{\neg\phi\}$.*

*Proof:* By Lemma 5.4 there is at most one $\mathcal{P}$-preferred$^L$ extension for each member of $\mathcal{C}$, and because the defaults in members of $\mathcal{C}$ are normal, there is at least one extension for each member of $\mathcal{C}$ by Theorem 3.1 in (Reiter, 1980). Therefore there is exactly one $\mathcal{P}$-preferred$^L$ extension for each member of $\mathcal{C}$. A decision procedure for the class of default theories stated in the theorem is given in Figure 8. The procedure computes a set $E$ that consists of $W$ and the conclusions of the generating defaults of the extension $Cn(E)$ of $\langle D, W \rangle$, and returns *true* if and only if $E \models \phi$.

Let ${:}\phi_1/\phi_1, \ldots, {:}\phi_n/\phi_n$ be the ordering $\mathcal{P}$ of defaults in $D$. If $W$ is inconsistent the procedure returns *true*. In this case $Cn(W) = \mathcal{L}$ is the only extension of $\langle D, W \rangle$ which agrees with the statement of the lemma. Otherwise extensions of $\langle D, W \rangle$ are consistent. Define $D_i = \{ {:}\phi_1/\phi_1, \ldots, {:}\phi_i/\phi_i \}$ and $\Pi_i = \{\phi_1, \ldots, \phi_i\}$ for all $i \in \{0, \ldots, n\}$. The correctness proof is by induction on $i$ and we obtain the $\mathcal{P}$-preferredness$^L$ of $Cn(E)$ as the case $i = n$.

*Induction hypothesis*: for all $j \in \{0, \ldots, i\}$, if $W$ is consistent, $E_j$ is a maximal consistent subset of $\Pi_j \cup W$ such that $W \subseteq E_j$, and $Cn(E_i)$ is a $\mathcal{P} \cap (D_i \times D_i)$-preferred$^L$ extension of $\langle D_i, W \rangle$.

The induction proof is straightforward. That the procedure runs in polynomial time on the size of $\langle D, W \rangle$ is also obvious. Thus any polynomial time subset $\mathcal{C}$ of classical propositional logic produces a tractable class of default theories. $\qquad\square$

Without priorities reasoning with prerequisite-free normal default rules and Horn clauses is NP-complete (Stillman, 1990). For normal defaults, tractability can be obtained also for defaults with prerequisites if suitable restrictions are imposed on the prerequisites, conclusions, and the objective parts of the default theories. Next we present such a class that is neither subsumed by nor subsumes the tractable prerequisite-free classes. The idea behind

```
PROCEDURE decide(D, W, P, φ)
    LET :φ₁/φ₁, :φ₂/φ₂, . . . , :φₙ/φₙ be the ordering P on D;
    E₀ := W;
    FOR i := 1 TO n DO
        IF Eᵢ₋₁ ∪ {φᵢ} is consistent (equivalently, Eᵢ₋₁ ⊭ ¬φᵢ)
        THEN Eᵢ := Eᵢ₋₁ ∪ {φᵢ}
    END;
    E := Eₙ;
    IF E ⊨ φ
    THEN RETURN true ELSE RETURN false
END
```

Figure 8: A decision procedure for a class of prerequisite-free normal default theories

the result is the same as the one used in the proof of Theorem 5.15. Corollary 5.18 indicates that the result cannot be generalized (assuming P$\neq$NP) to any of the classes higher in the Kautz and Selman hierarchy.

**Theorem 5.22** *Let $\mathcal{C}$ be the class of Horn default theories $\langle D, W \rangle$ where $W$ is a finite set of literals and defaults in $D$ are of the form $p_1 \wedge \cdots \wedge p_n : l/l$ where $l$ is a literal and $p_1, \ldots, p_n$ are propositional variables. For $\Delta \in \mathcal{C}$, negated Horn clauses $\phi$, and strict total orders $\mathcal{P}$ on the defaults in $\Delta$, the problem of testing $\Delta \models^L_\mathcal{P} \neg\phi$ is solvable in polynomial time on the size of $D \cup W \cup \{\neg\phi\}$.*

*Proof:* The body of the decision procedure is given in Figure 7 and its correctness proof is included in the proof of Theorem 5.15. Here we use the subprocedure *exists* in Figure 9. The procedure *exists* runs in polynomial time on the size of $D \cup W \cup A$ because the number of iterations in the repeat-until loop is at most $|D'| \leq |D|$ and logical consequence tests with propositional Horn clauses can be performed in polynomial time. Next we prove that the procedure *exists* called with arguments $(D, W, A)$ returns *true* if and only if there is an extension $E$ of $\Delta = \langle D, W \rangle$ such that $A \subseteq \mathrm{GD}(E, \Delta)$. The correctness proof and the algorithm are based on the idea that to derive the prerequisites of defaults in $A$, defaults in $D \backslash A$ with a negative conclusion are not needed.

First we show that *exists* returns true if and only if there is an extension $E'$ of $\Delta' = \langle D', W \rangle$ such that $A \subseteq \mathrm{GD}(E', \Delta')$, where $D' = A \cup \{\alpha{:}p/p \in D | \neg p \notin W, \alpha'{:}\neg p/\neg p \notin A, p \text{ is atomic}\}$. If $W \cup \{\pi|\alpha{:}\pi/\pi \in A\}$ is inconsistent and $A \neq \emptyset$, then there are no extensions that apply $A$, and hence it is correct to return *false*. So assume $W \cup \{\pi|\alpha{:}\pi/\pi \in A\}$ is consistent. Now the set $U = W \cup \{\pi|\alpha{:}\pi/\pi \in D'\}$ is consistent by definition of $D'$. Any extension $E'$ of $\Delta'$ satisfies $E' \subseteq Cn(U)$, and because all defaults in $D'$ are normal, no negation of a justification of a default in $D'$ is in $Cn(U)$. Therefore exactly those defaults are applied in extensions of $\Delta'$ for which the prerequisite is derivable. It is straightforward to show that the procedure computes the union $E$ of $W$ and the set of conclusions of the unique such set of defaults, and hence $Cn(E)$ is the unique extension of $\Delta'$. Finally, the procedure returns false if and only if $\alpha \notin E$ for some $\alpha{:}\pi/\pi \in A$. This is equivalent to the fact that $\delta \notin \mathrm{GD}(E, \Delta')$ for some $\delta \in A$, and hence $A \nsubseteq \mathrm{GD}(E, \Delta')$.

*PROCEDURE* exists$(D, W, A)$
  *IF* $A \neq \emptyset$ and $W \cup \{\pi | \alpha{:}\pi/\pi \in A\} \models \bot$ *THEN RETURN* false;
  $D' := A \cup \{\alpha{:}p/p \in D | \neg p \notin W, \alpha'{:}\neg p/\neg p \notin A, p \text{ is atomic}\};$
  $E := W;$
  *REPEAT*
    $E' := E;$
    *FOR EACH* $\alpha{:}\pi/\pi \in D'$ *DO*
      *IF* $E' \models \alpha$ *THEN* $E := E \cup \{\pi\}$
    *END*
  *UNTIL* $E = E';$
  *IF* $E \not\models \alpha$ for some $\alpha{:}\pi/\pi \in A$ *THEN RETURN* false
  *ELSE RETURN* true
*END*

Figure 9: A subprocedure of a decision procedure

Then we show that there is an extension $E'$ of $\Delta'$ such that $A \subseteq \mathrm{GD}(E', \Delta')$ if and only if there is an extension $E$ of $\Delta$ such that $A \subseteq \mathrm{GD}(E, \Delta)$. The "only if" direction directly follows from Theorem 3.2 in (Reiter, 1980) because $D' \subseteq D$ and $D$ is normal. Assume that there is an extension $E$ of $\Delta$ such that $A \subseteq \mathrm{GD}(E, \Delta)$. It is straightforward to show that $E$ is an extension of $\langle \mathrm{GD}(E, \Delta), W \rangle$. Next we remove defaults with negative conclusions that are in $D$ but not in $A$: an induction proof with Theorem 3.2 and the induction hypothesis $E_i'' = Cn(E_i \cap (\{\pi | \alpha{:}\pi/\pi \in A\} \cup \{p | \alpha{:}p/p \in \mathrm{GD}(E, \Delta), p \text{ is atomic}\} \cup W))$ shows that $E'' = Cn(\{\pi | \alpha{:}\pi/\pi \in A\} \cup \{p \in E | \alpha{:}p/p \in D, p \text{ is atomic}\} \cup W)$ is an extension of $\langle \mathrm{GD}(E, \Delta) \cap D', W \rangle$ and $A \subseteq \mathrm{GD}(E'', \langle \mathrm{GD}(E, \Delta) \cap D', W \rangle)$. By Theorem 3.2 in (Reiter, 1980), there is an extension $E'$ of $\Delta'$ such that $A \subseteq \mathrm{GD}(E', \Delta')$.

Therefore the procedure returns *true* if and only if there is an extension $E$ of $\langle D, W \rangle$ such that $A \subseteq \mathrm{GD}(E, \Delta)$.     □

The tractability result is related to the tractability of brave reasoning of the same class as shown by Lemma 6.4 in (Kautz & Selman, 1991).

### 5.2.4 Results for Arbitrary Priorities

The previous section restricts to the special case where priorities are a strict total ordering on the defaults. However, for instance in representing inheritance networks, two defaults sometimes need to have an equal (or incomparable) priority. Hence the possibility of tractable inference with less restricted priorities is of interest. The complexity results for unrestricted priorities are summarized in Table 6. Like in earlier sections, references to theorems are given in Table 7. The results in the previous sections as well as results on the complexity of cautious reasoning (Kautz & Selman, 1991) directly imply the intractability of many classes of reasoning with arbitrary priorities, because the former two are a special case of the latter. Classes of default theories for which the tractability question remains open are the prerequisite-free normal classes and the normal unary class with literals. For prerequisite-free normal unary theories with 1-literal clauses reasoning is tractable, but the remaining classes are sufficiently expressive to encode propositional satisfiability. It turns out that the tractability of lexicographic prioritized default logic coincides with the tractability of Reiter's default logic for all but one class. There may still be differences in the complexity of the intractable classes, for example Reiter's default logic could be in co-NP and lexicographic prioritized default logic could be $\Delta_2^p$-hard. We have not analyzed the intractable classes in more detail.

| | class of default theories | complexity when clauses in W are | | |
|---|---|---|---|---|
| | | Horn | 2-literal | 1-literal |
| 1 | disjunction-free | co-NP-hard | co-NP-hard | co-NP-hard |
| 2 | unary | co-NP-hard | co-NP-hard | co-NP-hard |
| 3 | disjunction-free ordered | co-NP-hard | co-NP-hard | co-NP-hard |
| 4 | ordered unary | co-NP-hard | co-NP-hard | co-NP-hard |
| 5 | disjunction-free normal | co-NP-hard | co-NP-hard | co-NP-hard |
| 6 | Horn | co-NP-hard | co-NP-hard | co-NP-hard |
| 7 | normal unary | co-NP-hard | co-NP-hard | **co-NP-hard** |
| 8 | prerequisite-free | co-NP-hard | co-NP-hard | co-NP-hard |
| 9 | prerequisite-free ordered | co-NP-hard | co-NP-hard | co-NP-hard |
| 10 | prerequisite-free unary | co-NP-hard | co-NP-hard | co-NP-hard |
| 11 | prerequisite-free ordered unary | co-NP-hard | co-NP-hard | co-NP-hard |
| 12 | prerequisite-free normal | co-NP-hard | co-NP-hard | co-NP-hard |
| 13 | prerequisite-free normal unary | co-NP-hard | co-NP-hard | PTIME |
| 14 | prerequisite-free positive normal unary | co-NP-hard | co-NP-hard | PTIME |

Table 6: Complexity of the consequence relation $\models^L$ with arbitrary priorities

| | class of default theories | reference | | |
|---|---|---|---|---|
| | | Horn | 2-literal | 1-literal |
| 1 | disjunction-free | K&S | K&S | K&S |
| 2 | unary | K&S | K&S | K&S |
| 3 | disjunction-free ordered | K&S | K&S | K&S |
| 4 | ordered unary | K&S | K&S | K&S |
| 5 | disjunction-free normal | K&S | K&S | K&S |
| 6 | Horn | K&S | K&S | K&S |
| 7 | normal unary | 14 $\subseteq$ | 14 $\subseteq$ | T5.24 |
| 8 | prerequisite-free | 14 $\subseteq$ | 14 $\subseteq$ | 11 $\subseteq$ |
| 9 | prerequisite-free ordered | 14 $\subseteq$ | 14 $\subseteq$ | 11 $\subseteq$ |
| 10 | prerequisite-free unary | 14 $\subseteq$ | 14 $\subseteq$ | 11 $\subseteq$ |
| 11 | prerequisite-free ordered unary | 14 $\subseteq$ | 14 $\subseteq$ | T4.14 |
| 12 | prerequisite-free normal | 14 $\subseteq$ | 14 $\subseteq$ | T4.13 |
| 13 | prerequisite-free normal unary | 14 $\subseteq$ | 14 $\subseteq$ | T5.23 |
| 14 | prerequisite-free positive normal unary | T4.12 | T4.12 | $\subseteq$ 13 |

Table 7: References to theorems on the complexity of $\models^L$ with arbitrary priorities

**Theorem 5.23** *Let $\mathcal{C}$ be the class of default theories $\Delta = \langle D, W \rangle$ where $W$ is a set of literals and $D$ consists of defaults of the form $:l/l$ where $l$ is a literal. Let $\mathcal{P}$ be a strict partial order on $D$ and let $l$ be a literal. Testing $\Delta \models^L_{\mathcal{P}} l$ can be done in polynomial time on the size of $\Delta$ and $\mathcal{P}$ and $l$.*

*Proof:* The algorithm in Figure 1 tests $\langle D, W \rangle \models^L_{\mathcal{P}} l$. The correctness of the algorithm is as follows. We analyze the *if*-statements in sequence. In each case we may use the negations of the assumptions of the previous cases. For the first four statements the proof is like the proof of Theorem 4.11 as no priorities are involved. 5. Assume that $:l/l\mathcal{P}:\bar{l}/\bar{l}$. Now $:l/l\mathcal{T}:\bar{l}/\bar{l}$ for all strict total orders $\mathcal{T}$ such that $\mathcal{P} \subseteq \mathcal{T}$. We obtain the unique extension with the $\Delta, \mathcal{P}$-ordering $\mathcal{T}$ by the algorithm given in Figure 8 and proven correct in Theorem 5.21. Obviously, $l$ is in that extension, and consequently in all $\mathcal{P}$-preferred$^L$ extensions. Hence it is correct to return *true*. 6. In the remaining case not $:l/l\mathcal{P}:\bar{l}/\bar{l}$. Hence there is a strict total ordering on $D$ such that $\mathcal{P} \subseteq \mathcal{T}$ and $:\bar{l}/\bar{l}\mathcal{T}:l/l$. An argument similar to the one in the previous case shows that there is a $\mathcal{P}$-preferred$^L$ extension with the $\Delta, \mathcal{P}$-ordering $\mathcal{T}$ that contains $\bar{l}$ and therefore does not contain $l$. Hence it is correct to return *false*.

Therefore the algorithm returns true if and only if $l$ is in all $\mathcal{P}$-preferred$^L$ extensions of $\Delta$. Clearly, the algorithm runs in polynomial time. □

Without priorities, cautious and brave reasoning for normal unary theories and 1-literal clauses is tractable (Kautz & Selman, 1991). We show that priorities increase the expressivity sufficiently to make this class intractable.

**Theorem 5.24** *The problem of testing whether a literal $l$ belongs to all $\mathcal{P}$-preferred$^L$ extensions of $\Delta$, where $\Delta$ is a normal unary default theory and $\mathcal{P}$ is a strict partial order on defaults in $\Delta$, is co-NP-hard.*

*Proof:* The proof is by reduction from propositional satisfiability to the complement of the problem. Let $C = \{c_1, \ldots, c_m\}$ be a set of propositional clauses and $P$ the set of propositional variables occurring in $C$. Let $N$ be an injective function that maps each clause $c \in C$ to a propositional variable $n = N(c)$ such that $n \notin P$. Define the default theory $\Delta = \langle D, \emptyset \rangle$ and priorities on $D$ as in the proof of Theorem 4.15. We claim that the set of clauses $C$ is satisfiable if and only if $\Delta \not\models^L_{\mathcal{P}}$ *false*; that is, there is a $\mathcal{P}$-preferred$^L$ extension of $\Delta$ that does not contain *false*. In the proof we refer to the consistency of extensions of $\langle D, W \rangle$ which is by the consistency of $W$ and the fact that defaults in $D$ have justifications (Corollary 2.2 by Reiter (1980)).

($\Rightarrow$) Assume that there is a model $M$ such that $M \models C$. We show that there is a $\mathcal{P}$-preferred$^L$ extension $E$ of $\Delta$ such that *false* $\notin E$. Let $E = Cn(\{p \in P | M \models p\} \cup \{p' | p \in P, M \not\models p\} \cup \{\neg p | p \in P, M \not\models p\} \cup \{\neg p' | p \in P, M \models p\} \cup \{n | c \in C, n = N(c)\} \cup \{\neg n' | c \in C, n = N(c)\})$. It is straightforward to verify that $E$ is an extension of $\langle D, \emptyset \rangle$. Let $\mathcal{T}$ be any strict total order on $D$ such that $\mathcal{P} \subseteq \mathcal{T}$ and for all $\{\delta, \delta'\} \subseteq D_1$ and all $\{\delta, \delta'\} \subseteq D_3$, $\delta \mathcal{T} \delta'$ if appl$(\delta, E)$ and not appl$(\delta', E)$. We show that $\mathcal{T}$ is a $\langle D, \emptyset \rangle, \mathcal{P}$-ordering for $E$. Let $E'$ be any extension of $\langle D, \emptyset \rangle$ such that there is $\delta \in D$ such that appl$(\delta, E', E)$. We show that there is $\delta' \in D$ such that appl$(\delta', E, E')$.

Assume that $\delta \in D_1$. Now $\delta = :l/l$ for some literal $l$ and $l \in E'$ and $\bar{l} \in E$. Because $E'$ is consistent, $\bar{l} \notin E'$. Hence appl$(:\bar{l}/\bar{l}, E, E')$. By definition $:\bar{l}/\bar{l}\mathcal{T}:l/l$.

Assume that $\delta \in D_2$ and $\delta = :p'/p'$ for some $p \in P$. Now by definition $p \in E$ and appl($p{:}\neg p'/\neg p', E$). Because $E'$ is consistent, $\neg p' \notin E'$ and hence appl($p{:}\neg p'/\neg p', E, E'$). By definition $p{:}\neg p'/\neg p'\mathcal{P}{:}p'/p'$.

Assume that $\delta \in D_2$ and $\delta = p{:}\neg p'/\neg p'$ for some $p \in P$. Because $p \in E'$, appl($:p/p, E'$). Because not appl($p{:}\neg p'/\neg p', E$) by definition $p \notin E$, and hence not appl($:p/p, E$). Hence $\neg p \in E$ and appl($:\neg p/\neg p, E$). Because $E$ is consistent, $\neg p \notin E$ and appl($:\neg p/\neg p, E, E'$). By definition $:\neg p/\neg p\mathcal{P}p{:}\neg p'/\neg p'$.

Assume that $\delta = p{:}n/n$. Clearly $p \notin E$, appl($:\neg p/\neg p, E, E'$) and $:\neg p/\neg p\mathcal{T}p{:}n/n$. Proof for $\delta = p'{:}n/n$ is similar.

Assume that $\delta = :n'/n'$. Hence appl($n{:}\neg n'/\neg n', E, E'$). By definition $n{:}\neg n'/\neg n'\mathcal{T}{:}n'/n'$. Proof for $\delta = n{:}\neg n'/\neg n'$ is similar.

Assume that $\delta = n'{:}false/false$. Hence appl($:n'/n', E', E$) and appl($:\neg n'/\neg n', E, E'$). By definition $:\neg n'/\neg n'\mathcal{T}n{:}false/false$.

This exhausts all $\delta \in D$. Therefore $\mathcal{T}$ is a $\langle D, \emptyset \rangle$, $\mathcal{P}$-ordering for $E$ and $E$ is a $\mathcal{P}$-preferred[L] extension of $\langle D, \emptyset \rangle$.

($\Leftarrow$) Assume that $E$ is a $\mathcal{P}$-preferred[L] extension of $\Delta$ such that $false \notin E$. We show that there is a model $M$ such that $M \models C$. Define the model as $M \models p$ iff $p \in E$, for all $p \in P$. Let $c = \{l_1, \dots, l_n\}$ be any clause in $C$. We show that $M \models c$. Let $n = N(c)$. Because $false \notin E$, not appl($n'{:}false/false, E$). Because $\neg false$ is not in any conclusion of a default in $D$, $n' \notin E$. Because not appl($:n'/n', E$), $\neg n' \in E$. Hence appl($n{:}\neg n'/\neg n', E$). Hence $n \in E$. Because $E$ is consistent, appl($q{:}n/n, E$) for some $q \in \{p, p'\}$ where $p \in c$ or $\neg p \in c$. If $q = p$, then $p' \notin E$ because of the following. Assume that $p' \in E$. Let $E_{12} = Cn(E \cap (P \cup \{\neg p | p \in P\} \cup \{p' | p \in P\} \cup \{\neg p' | p \in P\}) \backslash \{p'\} \cup \{\neg p'\})$. It is easy to show that $E_{12}$ is an extension of $\langle D_1 \cup D_2, \emptyset \rangle$. By Theorem 3.2 in (Reiter, 1980) there is an extension $E'$ of $\langle D, \emptyset \rangle$ such that $\mathrm{GD}(E_{12}, \langle D_1 \cup D_2, \emptyset \rangle) \subseteq \mathrm{GD}(E', \langle D, \emptyset \rangle)$. Now appl($p{:}\neg p'/\neg p', E', E$) and there is no $\delta \in D$ such that appl($\delta, E, E'$) and $\delta\mathcal{T}p{:}\neg p'/\neg p'$ for some strict total order $\mathcal{T}$ such that $\mathcal{P} \subseteq \mathcal{T}$. This contradicts the $\mathcal{P}$-preferredness[L] of $E$, and hence it must be the case that $p' \notin E$. Hence by definition $M \models p$. If $q = p'$, then clearly $p' \in E$. Hence by definition $M \models \neg p$. Therefore $M \models c$. Because this holds for any clause $c \in C$, $M \models C$. $\square$

## 6. Related Work on Prioritized Default Reasoning

Marek and Truszczyński (1993) introduce a prioritized default logic that is similar to Brewka's (1994) logic. The existence of preferred extensions is not guaranteed in general, but for normal defaults it is. We believe that for normal default theories, the complexity of the Marek and Truszczyński logic coincides with the complexity of Brewka's logic. Delgrande and Schaub (1997) present a translation from prioritized default theories to unprioritized default theories, so that the extensions of the resulting theories obey the priorities. Their translation can be performed in polynomial time. Buccafurri et al. (1998) present a knowledge representation language that extends logic programs with priorities, classical negation, and disjunction. They claim that brave reasoning for their language is in general $\Sigma_2^p$-complete and without disjunction and classical negation it is polynomial time.

Brewka and Eiter (1998) present the notion of *preferred answer sets* for extended logic programs. Their definition diverges from earlier work in that even with a total ordering on the rules there may be more than one preferred answer set, and preferred answer sets do not always exists for a given program even if answer sets do. Brewka and Eiter show that testing the membership of a literal in all preferred answer sets of a program is co-NP-hard when the rules are totally ordered. Their motivation for introducing preferred answer sets is that earlier accounts of priorities in logic programs and in default logic do not fulfill two principles identified by them. Principle I is violated by the prioritized default logics that are based on the semiconstructive definition of extensions (Baader & Hollunder, 1995; Brewka, 1994; Marek & Truszczyński, 1993), but not by the lexicographic prioritized default logic discussed in Section 5. Principle II can be paraphrased as follows. Let $E$ be a $\mathcal{P}$-preferred extension of $\langle D, W \rangle$ and $\alpha{:}\beta/\gamma$ a default such that $\alpha \notin E$. Then $E$ is a $\mathcal{P}'$-preferred extension of $\langle D \cup \{\alpha{:}\beta/\gamma\}, W \rangle$ for all $\mathcal{P}'$ such that $\mathcal{P}' \cap (D \times D) = \mathcal{P}$. Brewka and Eiter show that the prioritized default logic investigated in Section 5 violates this principle. However, Principle II is not violated by a closely related prioritized default logic that replaces application with non-defeat (Rintanen, 1999). A default $\alpha{:}\beta/\gamma$ is *defeated in* $E$ if $E \models \alpha \wedge \neg\beta$. Therefore it is not in general the case that lexicographic definitions of preferredness would violate Principle II. However, we believe that Brewka and Eiter would nevertheless find this definition of preferred extensions counterintuitive.

**Example 6.1** Let $D = \{a{:}b/b, {:}a/a, {:}\neg a/\neg a\}$ and $W = \{\neg b\}$. Let $\mathcal{T}$ be a strict total order on $D$ such that $a{:}b/b\,\mathcal{T}{:}a/a$ and ${:}a/a\,\mathcal{T}{:}\neg a/\neg a$. The default theory $\langle D, W \rangle$ has two extensions, $E_1 = Cn(\{a, \neg b\})$ and $E_2 = Cn(\{\neg a, \neg b\})$. The extension $E_1$ defeats $a{:}b/b$ and ${:}\neg a/\neg a$, and $E_2$ defeats ${:}a/a$. Because $E_1$ defeats the highest priority default $a{:}b/b$ and $E_2$ does not, only $E_2$ is a $\mathcal{T}$-preferred extension of $\langle D, W \rangle$. □

We believe that in this example, Brewka and Eiter would want $E_1$ to be the only $\mathcal{T}$-preferred extension because the only conflict is between ${:}a/a$ and ${:}\neg a/\neg a$, and the priority of ${:}a/a$ is higher. It seems that the question at hand is the meaning of priorities in cases where defaults that are less directly derivable have a higher or equal priority. Lexicographic comparison gives one meaning, a very natural one in our opinion, but Brewka and Eiter seem to want to ignore the higher priority. Instead of addressing Brewka and Eiter's concern by devising new definitions of preferred extensions, it could be addressed by not using priorities that produce unintended results. If a restriction to defaults with literal prerequisites and conclusions is made, this could be done simply by requiring that $l{:}m/n\,\mathcal{P}l'{:}m'/n'$ whenever $n = l'$. Priorities used in translating inheritance networks to prioritized default logic fulfill this requirement (Rintanen, 1999).

# 7. Conclusions

We have presented a thorough analysis of the complexity of three versions of prioritized default logic, giving results that place these logics in the polynomial hierarchy, and analyzing the question of intractability versus tractability for syntactically restricted classes of default theories.

The main results place the propositional variants of three general formalizations of prioritized default reasoning, the logics by Brewka (1994) and Baader and Hollunder (1995) that are based on the semiconstructive definition of extensions, and a formalization that is based on lexicographic comparison, on the lower levels of the polynomial hierarchy. As the first two formalizations closely resemble each other, it is not surprising that polynomial time translations between the decision problems of these two formalizations exist. There are also polynomial time translations to and from Reiter's default logic. The third formalization, that uses lexicographic comparison to select the preferred extensions, is not reducible to Reiter's default logic in polynomial time (under the standard complexity-theoretic assumptions.)

An analysis of the complexity of the decision problems in syntactically restricted cases, following earlier work by Kautz and Selman (1991) and Stillman (1990), identifies the effect of priorities on the boundary between tractability and intractability in the prioritized versions of the decision problems. With priorities that totally order the defaults and classical propositional reasoning that is tractable, for example with Horn clauses or 2-literal clauses, reasoning in Brewka's and in Baader and Hollunder's logics is polynomial time. For the formalization of prioritized default reasoning that uses lexicographic comparison, the same assumptions yield tractable reasoning only for the so-called Horn defaults and background theories without disjunction, as well as normal defaults without prerequisites and disjunction. When arbitrary priorities are allowed, in all three logics reasoning is tractable only when defaults are of the form $:l/l$ for literals $l$ and the background theories are sets of literals.

## Acknowledgements

This research was mainly carried out at the Helsinki University of Technology and finished at the University of Ulm while funded by the SFB 527 of the Deutsche Forschungsgemeinschaft. We gratefully acknowledge the generous support of the Finnish Cultural Foundation and the Finnish Academy of Science and Letters that helped completing this work.

We thank the four anonymous reviewers for pointing out some errors and for advice on restructuring the article and on including references to related work in Section 6. Holger Pfeifer provided valuable assistance in proof-reading.